\documentclass[preprint,12pt]{elsarticle}
\usepackage{amssymb}
\usepackage{epstopdf}
\usepackage[figurename=Fig.]{caption}

\journal{Pattern Recognition}

\makeatletter
\def\ps@pprintTitle{%
	\let\@oddhead\@empty
	\let\@evenhead\@empty
	\def\@oddfoot{\reset@font\hfil\thepage\hfil}
	\let\@evenfoot\@oddfoot
}
\makeatother
\usepackage{graphicx}
\usepackage{amssymb}
\usepackage[cmex10]{amsmath}
\usepackage{epsfig}
\usepackage{float}
\usepackage{accents}
\usepackage{graphicx}
\usepackage{caption}
\usepackage{amsmath, amsthm, amssymb}
\usepackage[linesnumbered,ruled,vlined]{algorithm2e}
\usepackage{subfig}
\DeclareMathOperator*{\argmin}{arg\,min} 

\begin{document}
\captionsetup[figure]{labelfont={bf},name={Fig.},labelsep=period}
\captionsetup[table]{labelfont={bf},name={Table},labelsep=newline}
\begin{frontmatter}

\title{Vote-boosting ensembles}

\author{Maryam Sabzevari, Gonzalo Mart\'{\i}nez-Mu\~noz and Alberto Su\'arez}
\address{Universidad Aut\'onoma de Madrid, Escuela Polit\'ecnica Superior, Dpto. de Ingenier\'{i}a Inform\'atica, C/Francisco Tom\'as y Valiente, 11, 28049 Madrid, Spain }
\begin{abstract}
Vote-boosting is a sequential ensemble learning method in which the individual classifiers are built on different weighted versions of the training data. To build a new classifier, the weight of each training instance is determined in terms of the degree of disagreement among the current ensemble predictions for that instance. For low class-label noise levels, especially when simple base learners are used, emphasis should be made on instances for which the disagreement rate is high. When more flexible classifiers are used and as the noise level increases, the emphasis on these uncertain instances should be reduced. In fact, at sufficiently high levels of class-label noise, the focus should be on instances on which the ensemble classifiers agree. The optimal type of emphasis can be automatically determined using cross-validation. An extensive empirical analysis using the beta distribution as emphasis function illustrates that vote-boosting is an effective method to generate ensembles that are both accurate and robust.
\end{abstract}

		
\end{frontmatter}	
\section{Introduction}
In ensemble learning, the outputs of a collection of diverse classifiers are
combined to exploit their complementarity, in the expectation that the global
ensemble prediction be more accurate than the individual ones
\cite{dietterichmethod_2000_ensemble}. The complementarity of the classifiers is
either an indirect consequence of diversity, as in bagging
\cite{breiman_1996_bagging} and random forests \cite{breiman_2001_random}, or
can be explicitly favored by design, as in negative correlation learning
\cite{liu+yao_1999_ensemble} and boosting
\cite{schapire_1990_strength,freund+schapire_1997_decision,friedman++_2000_additive,schapire_2012_boosting}.
In this manuscript, we present vote-boosting, an ensemble learning method of the
latter type, in which the progressive focus on a particular training
instance depends on the degree of disagreement 
among the predictions of the ensemble classifiers. 
By contrast to standard boosting algorithms, 
the strength of this emphasis is independent 
of whether the instance is correctly or incorrectly classified. 
The optimal emphasis profile depends on the characteristics of the classification problem considered and on the complexity of the base learners. 
In problems with no or low levels of noise in the class labels of the 
training instances the appropriate focus is 
on instances on which the classifiers disagree (i.e. instances for which the ensemble prediction has a high degree of uncertainty). 
As the noise level increases, especially when more flexible base learners 
(e.g. unpruned CART or random trees) are used, 
the emphasis on uncertain instances should be reduced. In fact, 
in problems with sufficiently high levels of  class-label noise, 
the optimal strategy is to assign larger weights to instances on 
which the ensemble classifiers agree. 
In practice, the determination of the optimal emphasis 
strategy can be automatically made through parameter selection
(e.g. though cross-validation on the training data). 
The results of an extensive empirical will be used to 
illustrate that vote-boosting is an effective method 
to build accurate ensembles that are both 
accurate and robust to class-label noise.

The article is organized as follows: In Section \ref{sec:previous_work}, we
provide a review of ensemble methods that are related to the current proposal.
Vote-boosting is described in section \ref{sec:vote_boosting}.
In this section, we show that the ensemble construction algorithm can
be viewed as an optimization by gradient descent in the functional space of
linear combinations of hypothesis. In Section \ref{sec:experimental_results},
the properties and performance of vote-boosting ensembles are analyzed in an
extensive empirical evaluation on synthetic and real-world classification tasks
from different domains of application. Finally, the conclusions of this study
are summarized in Section \ref{sec:conclusions}.
		
\section{Previous Work} \label{sec:previous_work}

There are a wide variety of methods to build ensembles. In this work, we focus on
homogeneous ensembles, which are composed of predictors of the same type. Each
of the predictors in the ensemble is built from a training set
composed of labeled instances. Once the individual predictors
have been built, their outputs are combined to reach a global ensemble decision. 
A wide range of alternatives can be used to
carry out this combination \cite{tulyakov++_2008_review}. Nevertheless, simple
strategies, such as averaging real-valued outputs, or majority voting, if the
individual classifiers yield class-labels, are generally effective
\cite{giacinto+roli_2005_theoretical}.

Randomization techniques can be used to generate collections of diverse 
classifiers. The objective is to build predictors 
whose errors are as independent as possible. If these
errors of these classifiers are independent, they can be averaged out by the
combination process. An example of these types of ensembles is bagging
\cite{breiman_1996_bagging}. The individual classifiers in a bagging ensemble
are built by applying a fixed learning algorithm to independent bootstrap
samples drawn from the original training data. In class-switching ensembles,
each member is built using a perturbed versions of the original training set, in
which the class labels of a fraction of instances are modified at random
\cite{breiman_2000_randomizing,martinez++_2005_switching}. 
Alternatively, diverse classifiers can be
built by including some randomized steps in the learning algorithm
itself. For instance, in one of the earliest works on ensembles of neural
networks \cite{hansen+salamon_1990_neural_network_ensembles}, one takes advantage
of the presence of multiple local minima in the optimization process that is
used to tune the synaptic weights and averages over the predictions of the
neural networks that result from different initializations. Random forests
\cite{breiman_2001_random}, which are one of the most effective ensemble methods
\cite{fernandezDelgado++_2014_do}, are built using a combination of data
randomization and randomization in the learning algorithm: The ensemble
classifiers are random trees, which are trained on bootstrap replicates of the
original training dataset. Other effective classifiers of this type are rotation
forests \cite{rodriguez++_2006_rotation} and ensembles of extremely randomized
trees \cite{geurts++_2006_extremely}.

An alternative to simply generating diversity
is to explicitly aim to increase the complementarity 
of the enseble classifiers. An example of this strategy 
is Negative Correlation Learning \cite{liu+yao_1999_ensemble}. 
In this method, complementarity is favored by simultaneously training 
all the classifiers in the
ensemble: The parameters of the individual classifiers and the weights for the
combination of their outputs are determined globally by minimizing a cost
function that penalizes coincident predictions. 
Another example is boosting. Boosting originally refers to the 
problem of building a strong learner out of a
collection of weak learners; i.e. learners whose predictive accuracy is only
slightly better than random guessing
\cite{schapire_1990_strength,friedman++_2000_additive,schapire_2012_boosting}.
AdaBoost is one of the most widely used boosting algorithms
\cite{freund+schapire_1997_decision}. In AdaBoost an ensemble is grown by
incorporating classifiers that progressively focus on instances that are
misclassified by the previous classifiers in the sequence. The individual
classifiers are built by applying a learning algorithm that can 
handle individual instance weights. Alternatively, weighted resampling 
in the training set can be used. The first classifier is obtained by
assuming equal weights for all instances. The subsequent classifiers are built
using different emphasis on each of the training instances. Specifically, to
build the $t$-th classifier in the sequence, the weights of instances that are
misclassified by the most recent classifier in the ensemble are increased.
Correspondingly, the weights of the correctly classified instances are reduced.
The final prediction of the ensemble is determined by weighted majority voting.
The weight of an individual classifier in the final ensemble prediction depends
on the weighted accuracy of this classifier on the training set. The margin of
an instance is defined as the sum of weighted votes for the correct class minus
the sum of weighted votes for the most voted incorrect class. Therefore,
misclassified instances have negative margins. In AdaBoost, the evolution of the
weight of a particular instance is a monotonically decreasing function of its
margin \cite{freund++_1999_short}. 

AdaBoost is one of the most effective ensemble methods
\cite{dietterich_2000_experimental,macklin+opitz_2011_popular,fernandezDelgado++_2014_do}.
However, it is not robust to class-label noise
\cite{quinlan_1996_bagging,bauer++_1999_empirical,khoshgoftaar++_2011_comparing}.
Specifically, AdaBoost gives unduly high weights to noisy instances, whose class
labels are incorrect. There are numerous studies that address this excessive
sensitivity of AdaBoost to class-label noise
\cite{loureiro++_2004_outlier,abe++_2006_outlier,domingo+watanabe_2000_mAdaBoost,jiang_2001_is,verdejo++_2008_dynamically,shivaswamy+jebara_2011_variance,friedman++_2000_additive,ratsch++_1998_improvement,guo++_2002_norm,sun++_2004_two,sun++_2006_reducing,freund_2001_adaptive,freund_2009_more,cheamanunkul++_2014_nonconvex}.
A possible strategy is to identify and either remove noisy instances in the
training data, or correct their class-labels
\cite{loureiro++_2004_outlier,abe++_2006_outlier}. Another alternative is to
apply explicit or implicit regularization techniques to avoid assigning
excessive weight to a reduced group of instances
\cite{domingo+watanabe_2000_mAdaBoost,jiang_2001_is,verdejo++_2008_dynamically,mayhuaLopez++_2012_real,
shivaswamy+jebara_2011_variance}.
For instance, the logistic loss function employed in LogitBoost
\cite{friedman++_2000_additive} gives less emphasis to instances with large
negative margins than the exponential loss function used in AdaBoost. In
consequence, LogitBoost is generally more robust to class-label noise
\cite{mcdonald++_2003_empirical}. In other studies, penalty terms are used in the
cost function  to avoid focusing on outliers or on instances that are difficult
to classify
\cite{ratsch++_1998_improvement,guo++_2002_norm,sun++_2004_two,sun++_2006_reducing}.
It is possible to also use hybrid weighting methods that modulate the emphasis
on instances according to their distance to the decision boundary
\cite{verdejo++_2006_boosting,verdejo++_2008_dynamically,ahachad++_2015_neighborhood}.
Most boosting algorithms use convex loss functions. This has the advantage that
the resulting optimization problem can be solved efficiently using, for
instance, gradient descent. However, as shown in \cite{long+servedio_2010_random},
the generalization capacity of boosting variants that use convex loss
functions can be severely affected by class-label noise. Alternative non-convex
loss functions are used in BrownBoost and other robust boosting variants
\cite{freund_2001_adaptive,freund_2009_more,cheamanunkul++_2014_nonconvex}. In
these methods, the evolution of the weights is {\it not} a monotonic function of
the margin: instances with small negative margins (i.e. misclassified instances
that are close to the decision boundary) are assigned higher weights, as in
AdaBoost. However, instances whose margin is negative and large receive lower
weights. The rationale for using this type of emphasis is that instances in
regions with a large class overlap tend to have small margins. Focusing on these
instances is beneficial because the classification boundary can be modeled in
more detail. By contrast, large negative margins correspond to misclassified
instances that are far from the classification boundary. A robust boosting
algorithm should therefore avoid emphasizing these instances, which are likely
to be noisy.

In the next section we introduce vote-boosting, a novel 
boosting algorithm in which the weights of the instances
are determined in terms of the degree of agreement or 
disagreement among the predictions of the ensemble
members, irrespective of their actual class labels.
As illustrated by the results of
empirical evaluation presented
in section  \ref{sec:experimental_results}, 
the optimal type of emphasis 
(that is, whether the focus should be placed
on instances on which the classifiers disagree, 
or on instances on which they agree) can be determined
from the training data alone 
using, for instance, cross-validation.
Since the instance weights do not depend on whether the 
predictions by the ensemble classifiers
are correct, it is possible to avoid unduly  
emphasizing incorrectly classified instances that 
are outliers, which is one of the weaknesses of 
standard boosting algorithms, such as AdaBoost. 
In this manner, one can build accurate ensembles
that are robust to class-label nosie.

\section{Vote-boosting} \label{sec:vote_boosting}
 
Consider the problem of automatic induction of a classification system from
labeled training data. The original training set is composed of $N_{train}$
attribute class-label pairs $\mathcal{D}_{train} = \left\{ \left(\mathbf{x}_i,y_i \right)
\right\}_{i=1}^{N_{train}}$, where  ${\mathbf{x}} \in \mathcal{X}$. In this
article, we focus on binary classification tasks, in which $y \in \{-1,1\}$.
Problems with multiple classes can be addressed with a number of strategies,
such as the ones used in combination with AdaBoost for this purpose
\cite{schapire_2012_boosting}. 

Let  $\left\{f_{\tau}(\cdot) \right\}_{\tau=1}^t$ be a partially-grown ensemble
of size $t$. The $\tau$-th classifier in the ensemble is a function $f_{\tau} :
\mathbb{R}^D \rightarrow \{-1,1\}$ that maps a vector of attributes $
\mathbf{x}$ to a class label $f_{\tau}(\mathbf{x})$. This function is obtained
by applying a base learning algorithm to a training set, taking into account the
individual instance weights $\left\{w_i^{[\tau]} \right\}_{i=1}^{N_{train}}$.

To obtain the prediction of the ensemble, the predictions of the 
individual classifiers are aggregated by weighted averaging  
\begin{equation}
\label{eq:aggregation} F_t(\mathbf{x}) = \sum_{\tau =1}^t \alpha_{\tau}^{[t]}
f_{\tau}(\mathbf{x}), \quad F_t \in \left\{-1,1\right\}, 
\end{equation} 
where
$\alpha_{\tau}^{[t]} \ge 0$ is the weight of the prediction of the
$\tau$ classifier. These weights are normalized $\sum_{\tau = 1}^t \alpha_{\tau}
= 1$. In (unweighted) majority voting one assumes that all the predictions have
the same weight , $\alpha_{\tau}^{[t]} = 1 / t$.
This simple voting scheme provides good overall results. For
this reason, it will be used in our implementation. 
Based on this aggregated output,
the final prediction of the ensemble of size $t$ is 
\begin{equation}
H_t(\mathbf{x}) = sign \left(F_t(\mathbf{x}) \right). 
\end{equation} 
Assuming that $t$ is odd, $ H_t(\mathbf{x}) \in \left\{-1,1\right\}$. At this
stage of the ensemble construction process, instance $\mathbf{x}$ can be
characterized by $t_{+}(\mathbf{x})$ and $t_{-}(\mathbf{x}) = 	t -
t_{+}(\mathbf{x})$, the counts of positive and negative votes, respectively. The
fractions of votes in each class are 
\begin{equation} \pi^{[t]}_{\pm}(\mathbf{x}) =
\frac{t_{\pm}(\mathbf{x})}{t};  \quad  \pi^{[t]}_{+}(\mathbf{x}) +
\pi^{[t]}_{-}(\mathbf{x}) = 1. 
\end{equation} 
These values can be used to quantify the level of certainty of the ensemble
prediction. Values $\pi^{[t]}_{+}(\mathbf{x})$ close to $0$ or $1$ correspond to
instances for which the predictions of most ensemble classifiers coincide.
Instances whose classification by the ensemble is uncertain are characterized by
$\pi^{[t]}_{+}(\mathbf{x})$ close to $1/2$. In
contrast to standard boosting algorithms, in vote-boosting,
the instance weights depend on the degree of agreement or
disagreement among the predictions of the individual classifiers, not on whether
these predictions are erroneous. 

The pseudo-code of the proposed vote-boosting algorithm is presented in
\figurename~\ref{alg:voteboost}. 

\begin{algorithm}
\KwIn{
\\
	  $\quad  \mathcal{D}_{train} =
	  \left\{ \left(\mathbf{x}_i,y_i \right) \right\}_{i=1}^{N_{train}}, 
	          \ \mathbf{x}_i \in \mathcal{X}, \ y_i \in \{-1,1\}$  \\
		$\quad T $          \% Ensemble size \\
		$\quad \mathcal{L}$ \% Base learning algorithm\\
		$\quad g(p)$        \% Non-negative emphasis function ($ 0 \le p \le 1$) \\
\\
\\
	}
	$F_0(\cdot) \gets 0$ \\
	$t_+(\mathbf{x}_i) \gets 0$ $\forall i =1,..., N_{train} $ \\
  $w_i^{[1]} \gets \frac{1}{N_{train}}$ $\forall i =1,..., N_{train} $ 
\\ 
\For{$t \gets 1 $ \textbf{to} $T$}{
		$f_t(\cdot) \gets \mathcal{L}\left(\mathcal{D}_{train}, \mathbf{w}^{[t]}\right)$ \\
		$t_+(\mathbf{x}_i) \gets t_+(\mathbf{x}_i) + \mathbb{I}(f_t(\mathbf{x}_i)>0)$   
		$\forall i =1,..., N_{train} $  \\
		$\pi_+(\mathbf{x}_i) =  \frac{t_+(\mathbf{x}_i)+1}{t+2}$  $\forall i =1,...,
N_{train}$\\ 
		$w_i^{[t+1]} \gets g\left(\pi_+(\mathbf{x}_i)\right)$ $\forall i =1,..., N_{train} $  \\
		Normalize $\mathbf{w}^{[t]}$\\
		$F_t(\cdot) \gets F_{t-1}(\cdot) + f_t(\cdot)$\\
	}
\KwOut{$H_T(\cdot)=sign(F_T(\cdot))$}
\caption{Vote-boosting algorithm with resampling}
\label{alg:voteboost}
\end{algorithm}

The final ensemble is composed of $T$ classifiers, each of which is built by
applying the base learning algorithm $\mathcal{L}$ on the training data with
different sets of instance weights. The weights of the instances can be taken
into account using weighted resampling with replacement 
\begin{eqnarray}
\mathcal{D}_{train}^{[t]} & = & \text{Sample}\left(\mathcal{D}_{train}, \mathbf{w}^{[t]}\right) \nonumber \nonumber \\
f_t & \gets& \mathcal{L}\left(\mathcal{D}_{train}^{[t]}\right).
\end{eqnarray}

For the induction of the first ensemble classifier all instances are assigned
the same weight. These weights are updated at each iteration according to the
tally of votes: Assuming that instance $\mathbf{x}_i$ has received
$t_+\left(\mathbf{x}_i\right)$ votes for the positive class at the $t$-th
iteration, we use the Laplace estimator of the probability that a classifier in
the ensemble outputs a particular class prediction
\begin{equation} \label{eq:class_prediction_probability_estimate}
\pi_{\pm}(\mathbf{x}_i) =  \frac{t_{\pm}(\mathbf{x}_i)+1}{t+2}. 
\end{equation}
Finally the weights are updated  
\begin{equation}
w_i^{[t+1]} = \frac{1}{\mathcal{Z}_{t+1}}g\left(\pi_+(\mathbf{x}_i) \right),
\quad   \text{for} \;\; i =1,..., N_{train}, 
\end{equation}
where $g: [0,1] \rightarrow \mathbb{R}^+$ is an emphasis function, which 
is non-negative, and
\begin{equation}
\mathcal{Z}_{t+1} = \sum_{i=1}^{N_{train}} g\left(\pi_+(\mathbf{x}_i) \right)  
\end{equation}
is a normalization constant. These weights are then used to build the following
classifier in the ensemble.
		
\begin{figure}[!t]
\centering
\includegraphics[width=\textwidth]{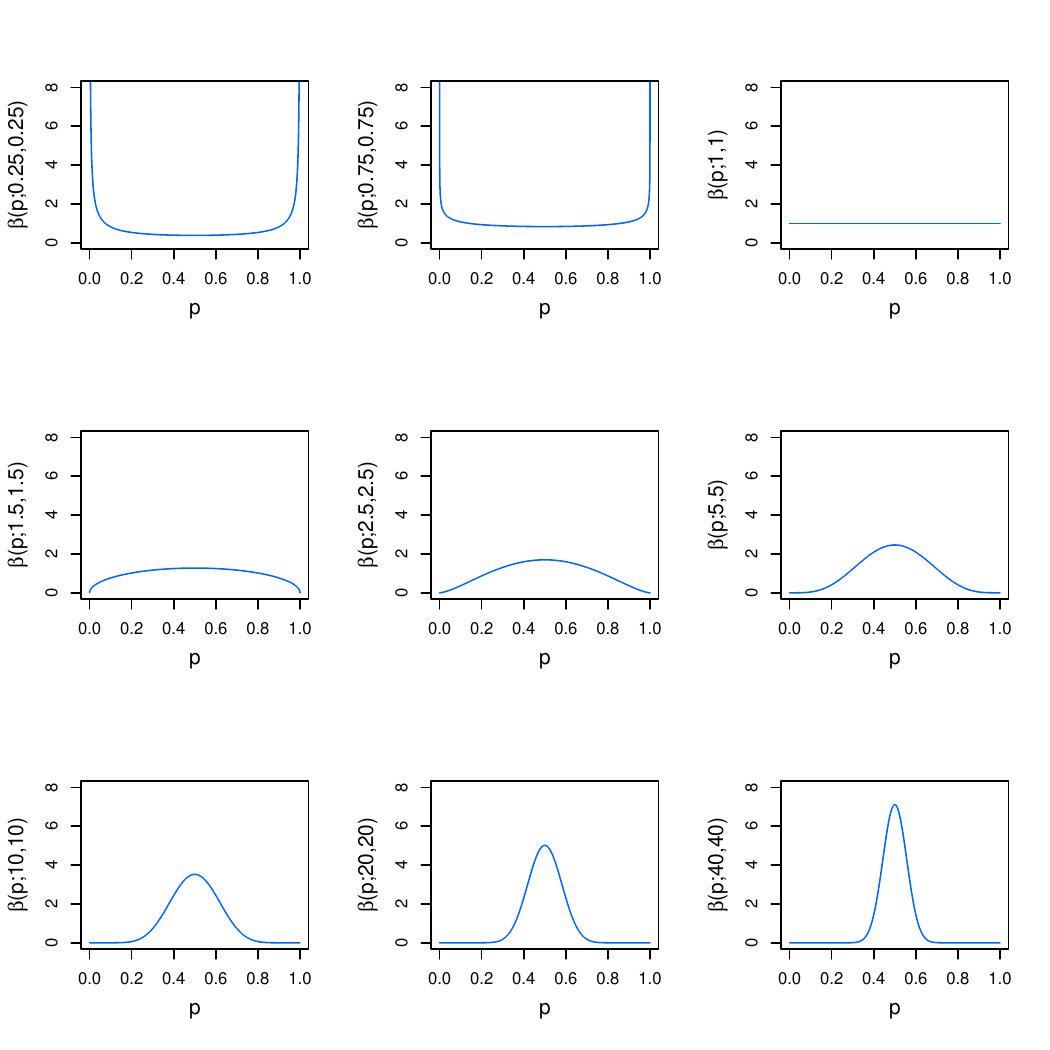}
\caption{Symmetric beta distribution with $a = b=[0.25,0.75,1,1.5,2.5,5,10,20,40]$}
\label{sym-beta}
\end{figure}

A natural choice for the emphasis function is the probability density of the
beta distribution with shape parameters  $a$, $b$ 
 \begin{equation}
g(p) \equiv \beta\left(p;a,b\right) = \frac{\Gamma(a+b)}{\Gamma(a)\Gamma(b)} p^{a-1} (1-p)^{b-1}, \quad  0 \le p \le 1.
\end{equation}
For this particular emphasis function, the weights at the
$(t+1)$-th iteration are updated according to 
\begin{equation} \label{eq:beta_weights}
w_i^{[t+1]} = \frac{1}{Z_{t+1}}  \beta\left(\pi_+(\mathbf{x}_i);a,b\right)
\end{equation}
If the class distributions are not strongly imbalanced, the choice $a = b$, in
which the two classes are handled in a symmetrical manner, is generally
appropriate. In problems with a large class imbalance, 
an asymmetric choice of the emphasis function may be preferable.
In \figurename~\ref{sym-beta} the density profiles of the symmetric
beta distribution for different values of $a$=$b \in
\{0.25,0.75,1,1.5,2.5,5,10,20,40\}$ are shown. 
If $a=b=1$, the distribution is uniform. 
Therefore, all instances are given the same importance
(plot in the first row, third column of
\figurename~\ref{sym-beta}). In this case, 
the proposed algorithm is equivalent to bagging
\cite{breiman_1996_bagging}. 
For $a=b > 1$ the distribution becomes unimodal, with a
maximum at $0.5$. 
In this range, the higher the values of $ a = b$, the more
concentrated becomes the probability around the mode. In consequence, 
vote-boosting emphasizes uncertain
instances and reduces the importance of those instances on which most
classifiers agree. For simple 
classifiers, the emphasis on uncertain instances that one obtains 
is similar to the error-based emphasis of AdaBoost. The reason is
that uncertain instances are generally more difficult to classify and, in
consequence, are more likely to be incorrectly classified. 
In the regime $a=b < 1.0$, vote boosting progressively focuses 
on instances that are far from the classification boundary. 
As will be illustrated in the section on experiments, this strategy is
effective in complex or noisy problems, 
especially when the ensemble is composed of flexible classifiers,
because of its regularizing effects.

It is common, especially in the
first iterations of the algorithm, when the ensemble is still small, that all
the classifiers predict the same class label for some instances.  In such cases,
the fraction of positive votes is either $0$ or $1$. Except for $a = b = 1$, the
value of the beta distribution at these points is either zero or infinity. In
the case of zero density values, those instances would be assigned zero weight
in the next iteration of the algorithm. Thus, they 
would be effectively removed from the sample. 
In the other extreme, some instances would have
infinite weights. To avoid these evaluations of the beta
distribution at the boundaries of its support, 
the Laplace correction has
been used in the estimation of class prediction probabilities
(\ref{eq:class_prediction_probability_estimate}).

Finally, we note that vote-boosting 
can be used with base learners that achieve zero or low error rates in  
the training data. In such cases, the weights that 
AdaBoost assigns to the training instances are ill-defined. 
By contrast, even if the training error of a single ensemble classifier is zero
or close to zero, there can be disagreement among the individual predictions. 
Therefore, provided that the Laplace correction is
used in the estimation of class prediction probabilities,
the weights given by Eq. \ref{eq:beta_weights}, 
are always well defined.
This feature allows us to build vote-boosting ensembles
of unpruned CART or random trees, which, as illustrated by 
the results of Section \ref{sec:experimental_results} are both accurate 
and robust to class-label noise.

\subsection{A functional gradient interpretation of vote-boosting} 
Similarly to
other boosting methods, vote-boosting can be viewed as a gradient descent
algorithm in the hypothesis space of linear combinations of predictors
\cite{mason++_1999_boosting}. Consider an ensemble of $t$ predictors $\left\{
f_{\tau} \right\}_{\tau = 1}^t$. The global ensemble prediction on
instance $\mathbf{x}$ is of the form
\begin{equation}
H_t(\mathbf{x}) = sign [F_t (\mathbf{x})],
\end{equation}
where
\begin{equation}
F_t(\mathbf{x}) =  \frac{1}{t} \sum\limits_{\tau=1}^{t}  f_{\tau}(\mathbf{x}), \quad F_t(\mathbf{x}) \in [-1,1].
\end{equation}
The fraction of votes for the positive class can be expressed in terms of this quantity
\begin{equation}
\pi_{+}^{[t]}(\mathbf{x}) = \frac{1 + F_t(\mathbf{x})}{2}. 
\end{equation}

Consider the 
predictor $F: \mathbf{x} \in \mathcal{X} \rightarrow F(x) \in [-1,1]$. 
Let the cost functional for this predictor be 
\begin{equation}\label{costfunc1}
C[F] = \frac{1}{N_{train}} \sum\limits_{i=1}^{N_{train}} y_i c(F(\mathbf{x}_i)),
\end{equation}
where $c(z)$ is a monotonically non-increasing function of $ z \in [-1,1]$, 
such that $c(0) = 0$. 
 These properties ensure that when the prediction 
of $F$ for the $i$-th example
is class $-1$ (i.e. $F(\mathbf{x}_i) < 0$), then $c(F(\mathbf{x}_i)) > 0$.
Similarly, when the prediction  
is class $+1$ (i.e. $F(\mathbf{x}_i) > 0$), then $c(F(\mathbf{x}_i)) < 0$.
Therefore, when the prediction 
$F(\mathbf{x}_i)$ is incorrect (i.e.  $y_i F(\mathbf{x}_i) < 0$), the contribution  
to the cost functional $y_i c\left(F(\mathbf{x}_i)\right)$ is positive.
When the prediction is correct, the corresponding contribution is negative.
From these properties ones concludes that
$C[F]$ achieves its global minimum when the training error
is zero. Furthermore, this quantity increases with each incorrect prediction. 
In consequence, the minimizer of $C[F]$ also minimizes 
the error of predictor $F$ in the training set.

If $\left| F(\mathbf{x}_i) \right|$ is a measure
of how certain the prediction of $F$ for instance $\mathbf{x}_i$ is,
the value $\left| c(F(\mathbf{x}_i)) \right|$  provides also a 
measure of such certainty.  
The magnitude of 
contribution $y_i c\left(F_t(\mathbf{x}_i)\right)$ to the cost functional
increases with the margin of the prediction.
It is largest when $\left| F_t(\mathbf{x}_i) \right| = 1$; 
that is, when the confidence when all ensemble 
classifiers agree; namely, when  $\left| F_t(\mathbf{x}_i) \right| = 1$. 

In vote-boosting, the first classifier in the ensemble 
is built by assuming equal weights for all the instances
in the training set.
Then, the ensemble is grown in a sequential manner 
by incorporating to 
$\left\{f_{\tau} \right\}_{\tau = 1}^t$, the 
ensemble of size $t$,  the classifier that 
minimizes the value of cost functional for the 
enlarged ensemble
$\left\{ f_{\tau} \right\}_{\tau = 1}^t \cup \left\{ f_{t+1} \right\} $
\begin{equation}
f_{t+1} = \argmin_{f \in {\mathcal{F}}}  C\left[F_{t+1}^{[f]}\right], 
\end{equation}
where
\begin{equation}
F_{t+1}^{[f]}(\mathbf{x}) 
= \frac{1}{t+1} \sum\limits_{\tau=1}^{t}  f_{\tau}(\mathbf{x}) +
\frac{1}{t+1} f(\mathbf{x})
= F_{t}(\mathbf{x})  
+ \frac{1}{t+1} \left(  f(\mathbf{x}) - F_{t}(\mathbf{x}) \right).
\end{equation}

Assuming the change in the value of the cost functional when the ensemble
incorporates the new classifier $f$ is small
\begin{eqnarray} \label{eq:delta_C_F}
\delta C[F_t] 
& \equiv & C\left[F_{t+1}^{[f]}(\mathbf{x})\right] -  C\left[F_{t}(\mathbf{x})\right] 
\nonumber \\ 
&  & \hspace*{-2.2cm} =
\frac{1}{N_{train}} \sum\limits_{i=1}^{N_{train}} y_i 
\left[ c\left(
F_{t}(\mathbf{x}_i)  
+ \frac{1}{t+1} \left(  f(\mathbf{x}_i) - F_{t}(\mathbf{x}_i) \right)
\right) - 
c\left(F_t(\mathbf{x}_i)\right) \right] 
 \nonumber \\  
&  & \hspace*{-2.2cm} 
\approx \frac{1}{(t+1) N_{train}}
\left[  \sum\limits_{i=1}^{N_{train}} y_i c'(F_t(\mathbf{x}_i)) f(\mathbf{x}_i) 
- \sum\limits_{i=1}^{N_{train}} y_i c'(F_t(\mathbf{x}_i)) F_t(\mathbf{x}_i) \right],
\end{eqnarray}
to lowest order in the Taylor expansion.
The second term in the last expression does not depend on $f$. Therefore, to lowest
order, minimizing the cost functional is equivalent to minimizing
\begin{align*}
& \hspace*{-1cm} \sum\limits_{i=1}^{N_{train}} y_i c'(F_t(\mathbf{x}_i)) f(\mathbf{x}_i)  \\
& =   \sum\limits_{i: y_i = f(x_i)} c'(F_t(\mathbf{x}_i)) -
\sum\limits_{i: y_i \neq f(x_i)} c'(F_t(\mathbf{x}_i))   \\
& = \sum\limits_{i=1}^{N_{train}} c'(F_t(\mathbf{x}_i)) - 2
\sum\limits_{i: y_i  \neq f(x_i)} c'(F_t(\mathbf{x}_i))  \\
& = 2 \sum\limits_{j=1}^{N_{train}} c'(F_t(\mathbf{x}_j))  
\left( \frac{1}{2} - \sum\limits_{i: y_i \neq f(x_i)} w_i^{[t+1]} \right),
\end{align*}
where 
\begin{equation} \label{eq:instance_weights}
w_i^{[t+1]} = \frac{c'\left(F_t(\mathbf{x}_i) \right)}{\sum_{j=1}^{N_{train}} 
c'\left(F_t(\mathbf{x}_j) \right) }. 
\end{equation}
Since $c(z)$ is a monotonic non-increasing function, 
then $-c'(z)$ is non-negative,
and the values $\left\{w_i^{[t+1]} \right\}_{i=1}^{N_{train}}$
defined in Eq. (\ref{eq:instance_weights}) 
can be thought of as a set of instance weights. 
The denominator in
(\ref{eq:instance_weights}) ensures that these weights are normalized
\begin{equation}
\sum_{i=1}^{N_{train}} w_i^{[t+1]} = 1.
\end{equation}
Using this expression for the instance weights, Eq. (\ref{eq:delta_C_F}) becomes
\begin{equation}
\delta C[F_t]   
\propto \sum\limits_{i: y_i \neq f_{t+1}(x_i)} w_i^{[t+1]} - \sum\limits_{i: y_i \neq F_t(x_i)} w_i^{[t+1]}.
\end{equation}
Note that $\delta C[F_t] < 0$ only if the weighted training error of 
the newly built classifier is lower than the corresponding error for the 
ensemble.

Under these conditions, the $(t+1)$-th predictor in the ensemble is the minimizer
of the {\it weighted} training error 
\begin{equation}
f_{t+1} = \arg\min_{f \in \mathcal{F}} \sum\limits_{i: y_i \neq f(x_i)} w_i^{[t+1]},
\end{equation}
where $\mathcal{F}$ is the functional space of the base learners.
In contrast to standard boosting algorithms, the weights given 
by (\ref{eq:instance_weights}) depend only on
the ensemble predictions, irrespective of whether these predictions are correct.

Assuming that the function $c(z)$ in (\ref{costfunc1}) is bounded,
it is convenient to express it in terms of a 
cumulative distribution function $G(\pi)$ defined in the unit interval 
$\pi \in [0,1]$  
\begin{eqnarray}
c\left(F(\mathbf{x})\right) & = & 2 K \left[ G(1/2) - G\left(\frac{1 + F(\mathbf{x})}{2} \right) \right], 
\end{eqnarray}
where $K$ is a positive constant. 
Without loss of generality, this constant is set to one ($K=1$).
Because of the monotonicity of $G(p)$, when the prediction 
$F(\mathbf{x}_i)$ is incorrect (i.e.  $y_i F(\mathbf{x}_i) < 0$), the contribution  
to the cost functional $y_i c\left(F(\mathbf{x}_i)\right)$ is positive.
Assuming this form for $c\left(F(\mathbf{x})\right)$, its derivative is 
\begin{eqnarray}
c'\left(F(\mathbf{x})\right) & = & - g\left(\frac{1 + F(\mathbf{x})}{2} \right),
\end{eqnarray}
where $g(p) = G'(p)$ is the corresponding probability density, which is non-negative.
With these assumptions, the weights of the training instances are
\begin{equation} \label{eq:weights_g}
w_i^{[t+1]} = \frac{g\left(\pi_{+}^{[t]}(\mathbf{x}_i)\right)}{ \sum_{j=1}^{N_{train}} 
g\left(\pi_{+}^{[t]}(\mathbf{x}_j) \right)}, \quad i = 1,2,\ldots, N_{train}, 
\end{equation}
where the density $g(p)$ plays the role of an emphasis function. 
Note that this density is not symmetric around 0.5. 
In fact, asymmetries in the emphasis could be useful in classification 
problems with unbalanced classes. 
In contrast to most boosting algorithms, including AdaBoost, 
the weights given by Eq. (\ref{eq:weights_g}) do not depend on 
the actual class label of the instance. Therefore, it
does not seem possible to derive error bounds similar 
to those enunciated in Theorem 6 (e.g. Eq. (21)) 
of \cite{freund+schapire_1997_decision}. 
 
\section{Empirical evaluation} \label{sec:experimental_results}

In this section, we present the results of an empirical analysis of vote-boosting. 
Different sets of experiments have been performed to analyze the properties
ensembles built with this method and evaluate their accuracy in a wide
range of classification tasks from different areas of application. In these
experiments, the symmetric beta distribution has been used as the emphasis
function. 
A first set of experiments is carried out to investigate the relationship of
vote-boosting with bagging and AdaBoost. The results of these experiments 
illustrate that, when simple (e.g. decision stumps) or regularized learners
(e.g. puned CART trees) are used as base learners, 
vote-boosting performs an interpolation 
between bagging ($a = b =1.0$) and AdaBoost (high values of $a=b$).
In a second set of experiments, we investigate the 
behavior of vote-boosting composed of different classifiers. 
In particular, we compare the accuracies of vote boosting ensembles
composed of decision stumps, pruned CART trees, unpruned CART trees, 
and (unpruned) random trees. The best overall results
in terms of predictive accuracy are obtained with 
random trees. However, the differences with vote-booting ensembles
composed of pruned or unpruned CART trees are not statistically significant.
Finally, the accuracy of vote-boosting ensembles composed of 
random trees 
is compared with bagging, AdaBoost and random forest. 
From the results of this benchmarking exercise, 
we conclude that vote-boosting ensembles composed of
random trees achieve state-of-the-art classification accuracy rates, which are
comparable or superior to random forest and AdaBoost in the problems
investigated.
A final batch of experiments is carried out to analyze the differences
among the optimal emphasis profiles for different classification problems
using random trees as base learners. 
This analysis illustrates
that in problems with low levels of noise in the
class labels, new classifiers should focus on instances whose 
classification by the current ensemble is uncertain
By contrast, in problems with contaminated labels,
the optimal emphasis is to reduce the weights of such uncertain instances, 
which are likely to be noisy. 
	
\subsection{Vote-boosting as an interpolation between bagging and AdaBoost}

The objective of the experiments presented in this subsection is to analyze
how the behavior of vote-boosting ensembles 
composed of simple or regularized learners, such as decision
stumps, or pruned CART trees, changes when different levels
of emphasis on the uncertain training instances are considered. As discussed
earlier, when uniform emphasis is made, vote-boosting is equivalent to bagging.
In most of the problems analyzed,  when such simple base learners
are used, stronger emphasis on
uncertain instances (i.e. instances for which 
the degree of disagreement among the 
ensemble predictions is largest) 
results in a behavior that is similar to AdaBoost. In such 
cases, vote-boosting provides an interpolation between bagging and AdaBoost,
depending on the strength of the emphasis on uncertain instances. 

\begin{figure*}[!t]
\centering
\subfloat[]{\includegraphics[scale=0.25]{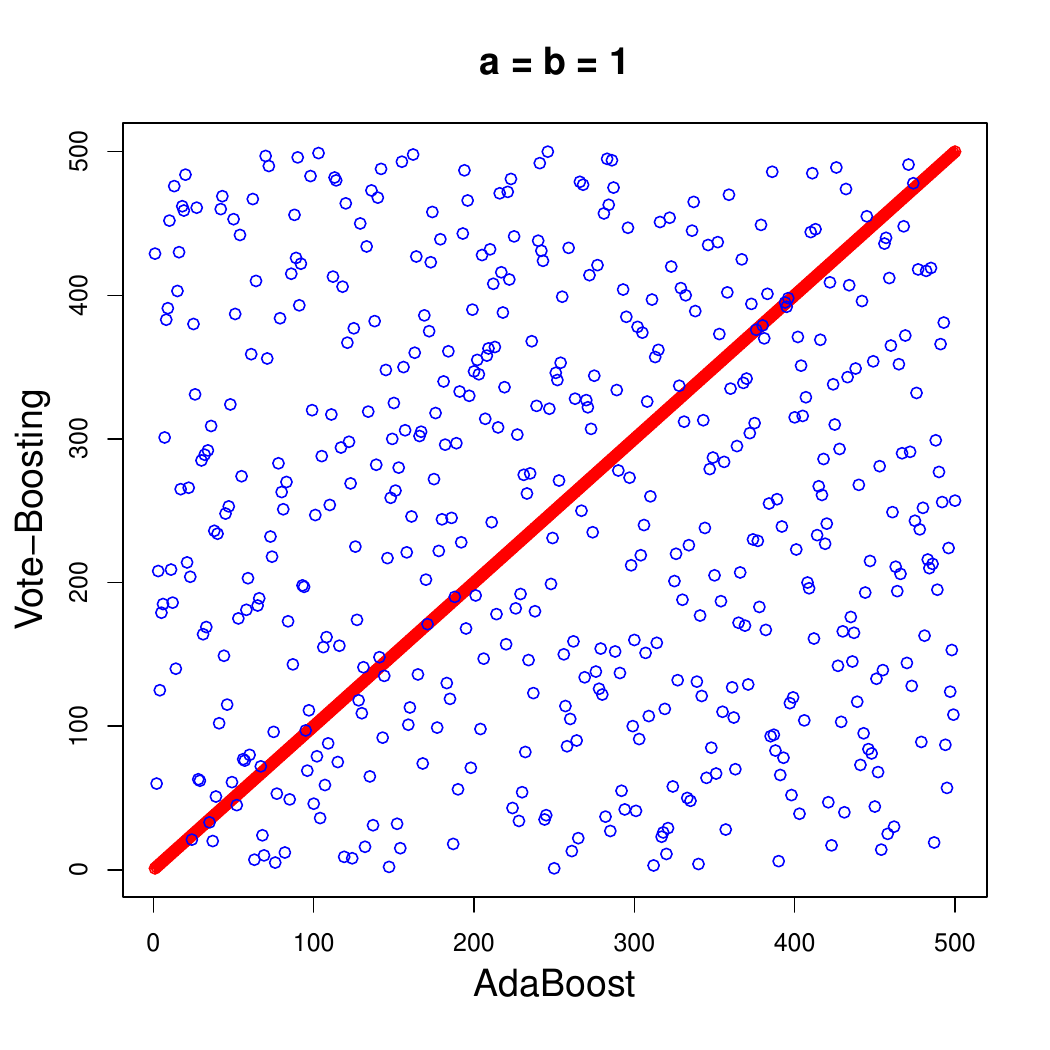}
	\label{fig1}}
\subfloat[]{\includegraphics[scale=0.25]{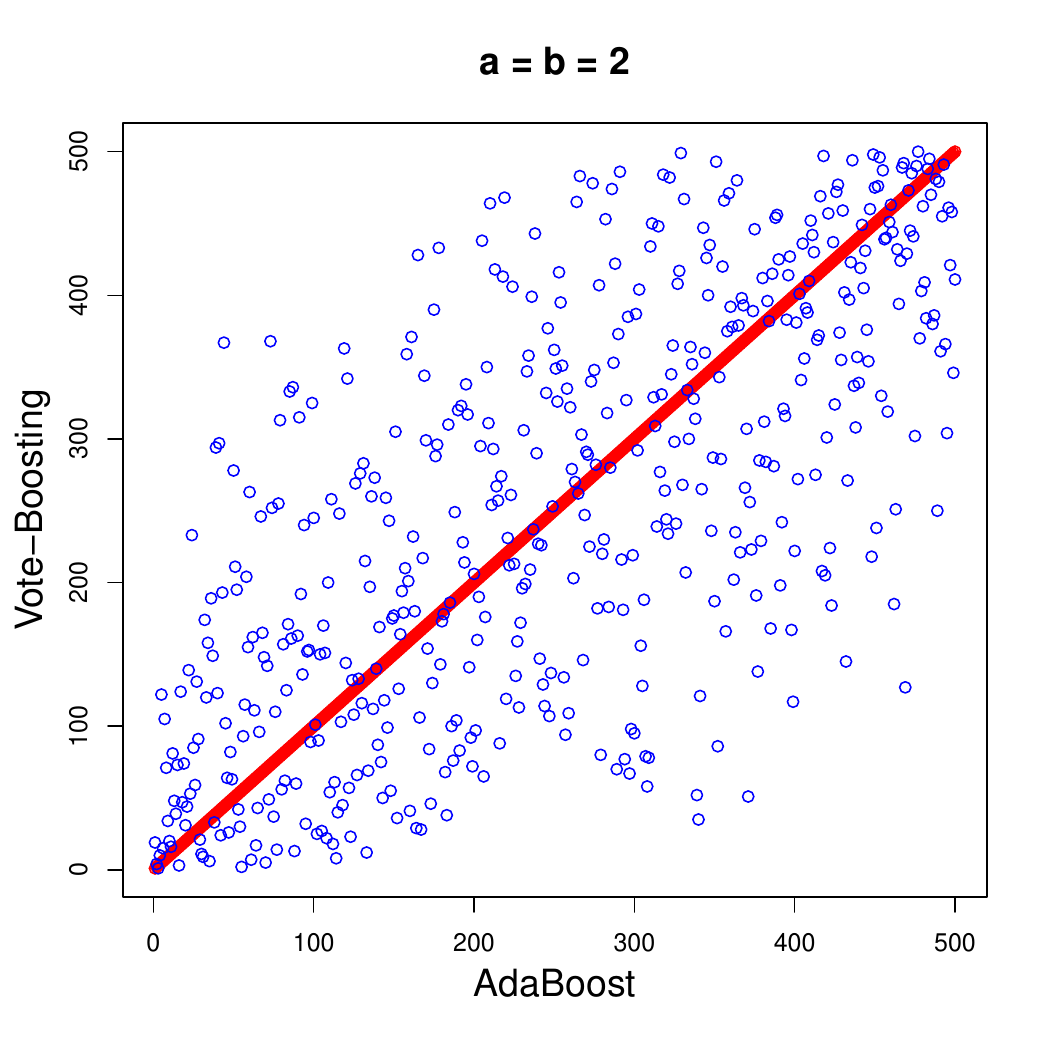}
	\label{fig3}}
\subfloat[]{\includegraphics[scale=0.25]{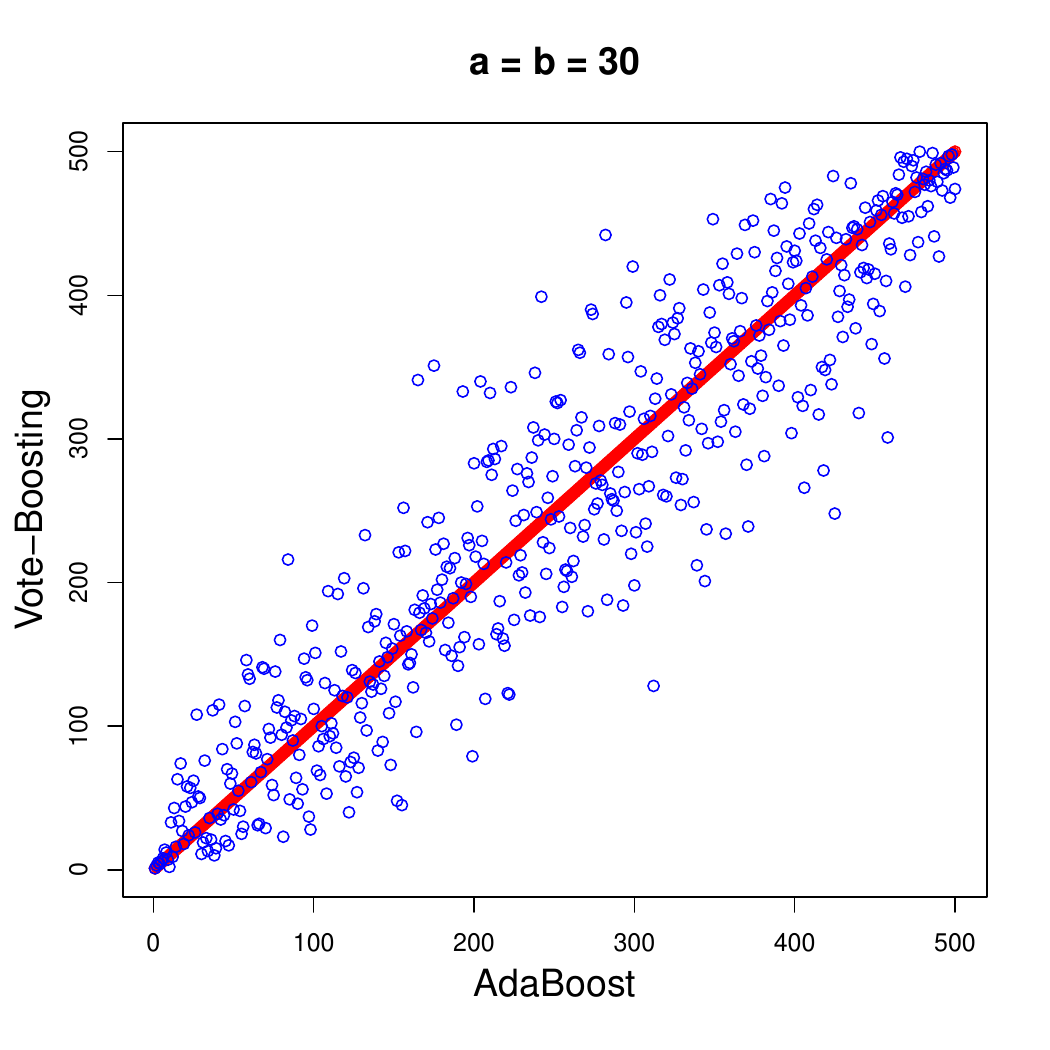}
	\label{fig4}}
\caption{Weight ranks of the training instances for vote-boosting and AdaBoost of decision stumps 
in {\it Twonorm} (a) $a = b = 1.0$, (b) $a = b = 2.0$, (d) $a = b = 30.0$}
\label{fig_won}
\end{figure*}

To investigate this relationship between vote-boosting and AdaBoost, we first
present the results of an experiment in the binary classification problem {\it
Twonorm} using decision stumps as base learners.  In {\it Twonorm}, instances
are drawn from two unit-variance Gaussians in $20$ dimensions whose means are $
(a,a,\ldots,a)$ for class $1$ and $(-a,-a,\ldots,-a)$ for class $2$, with $a = 2/
\sqrt{20}$ \cite{breiman_1996_bias}. This is not a trivial task for
decision stumps because, as individual classifiers, they can model only class
boundaries that are parallel to the axes. In the experiments performed, the
training set is composed of $500$ independently generated instances. Different
vote-boosting ensembles composed of $ 100 $ stumps were built using the
symmetric beta distribution for emphasis, with $a=b \in \left\{1.0, 2.0, 30.0
\right\}$. An AdaBoost ensemble composed of $100$ stumps was also built using
the same training data. The final weights given to the training instances in the
different ensembles were recorded and subsequently ranked. Ties were resolved by
randomizing the corresponding ranks. In \figurename~\ref{fig_won}, a scatter
plot of these ranks is shown where the ranks for AdaBoost are on the horizontal
axis and in the vertical axis for vote-boosting with $a = b = 1.0$ in (a), $a =
b = 2.0$ in  (b),  and $a = b = 30.0$ in (c). 
When $a = b = 1.0$ is used, all instances are assigned the same weight and no
correlations between the weight ranks given by vote-boosting and AdaBoost can be
observed. Therefore, the points corresponding to the training instances appear
uniformly distributed in \figurename~\ref{fig_won} (a). As the values of $a = b$
increase, points tend to cluster around the diagonal. This is a consequence
of the fact that the ranks of the weights given by both types of ensembles
become more similar. Comparing Figs. \ref{fig_won} (a), (b) and (c), it is
apparent that the correlations between the weight ranks become stronger as $a =
b$ increases. In particular for $a = b = 30.0$ vote-boosting and AdaBoost give
similar emphasis, even though the former does not make use of
class labels to decide whether an instance should be given more weight, whereas
the latter does. The reason for this coincident emphasis is that, in this simple
problem, the ensemble classifiers are more likely to disagree precisely in the
instances that are incorrectly classified.

\begin{figure*}[!t]
\centering
\subfloat[]{\includegraphics[scale=0.25]{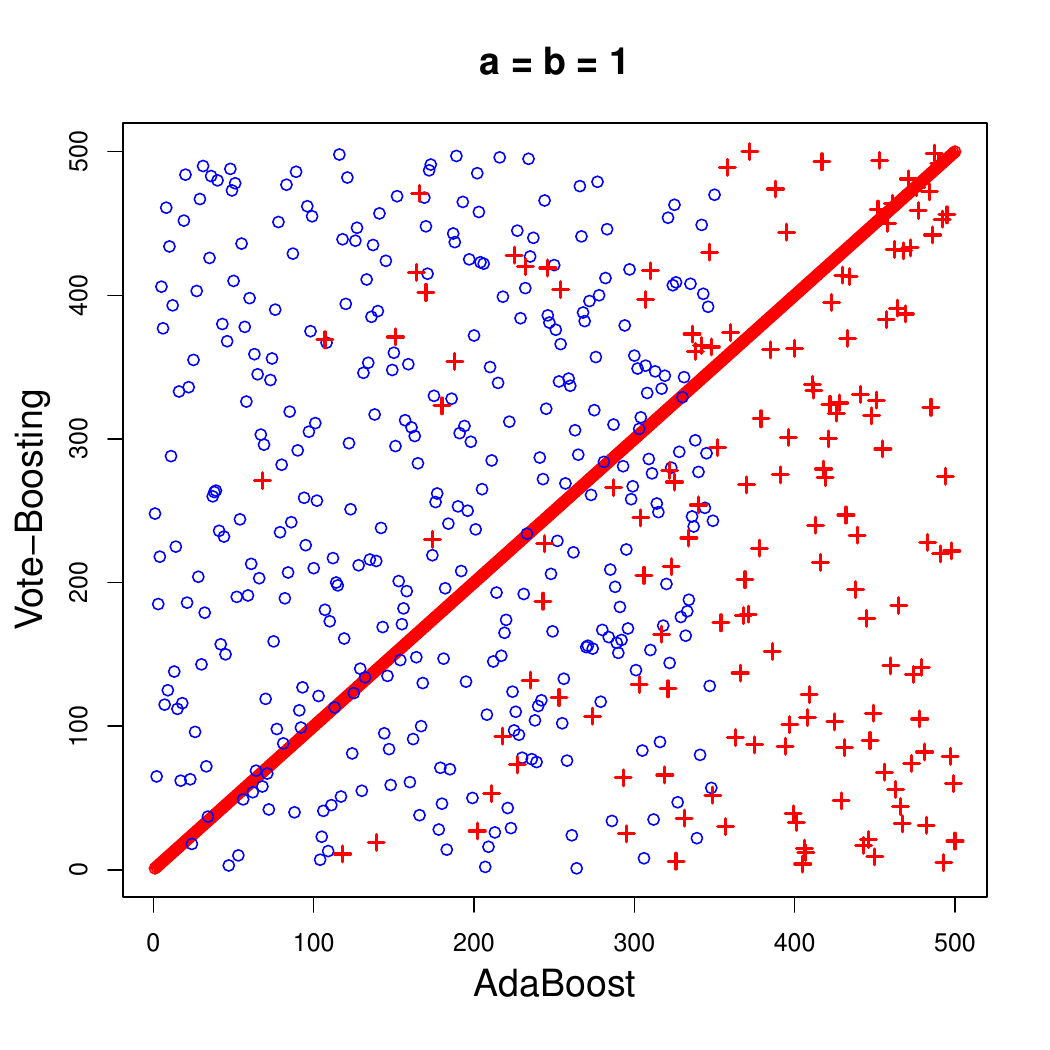}
	\label{fig11}}
\subfloat[]{\includegraphics[scale=0.25]{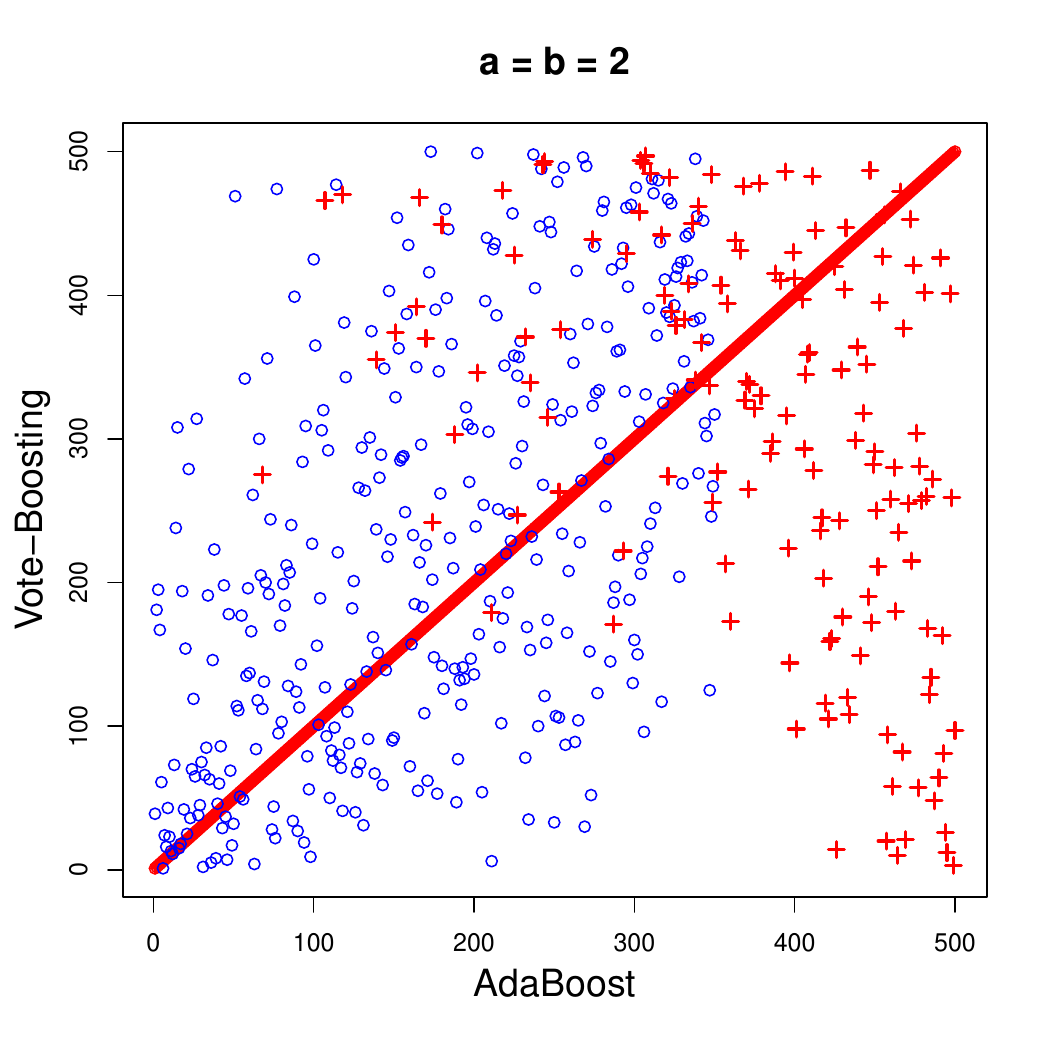}
	\label{fig13}}
\subfloat[]{\includegraphics[scale=0.25]{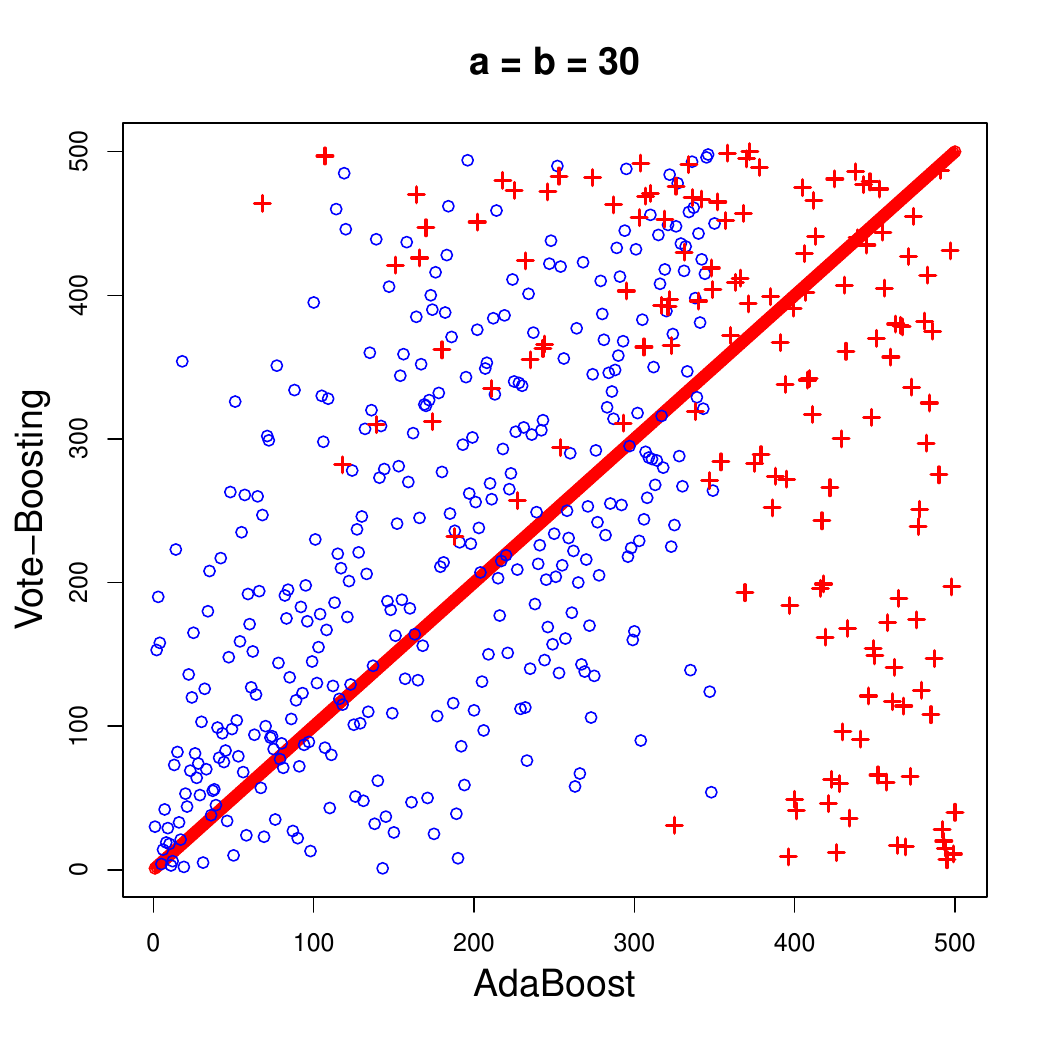}
	\label{fig14}}
\caption{Weight ranks of the training instances for vote-boosting and AdaBoost 
of decision stumps in {\it Twonorm} with $30\%$ class-label noise, (a) $a=b=1.0$, (b) $a=b=2.0$, (c) $a=b=30.0$ }
\label{fig_wn}
\end{figure*}

If class-label noise is injected in the problem, the weighting schemes
of AdaBoost and vote-boosting become different: Vote-boosting maintains the
focus on instances in the boundary region, in which classes overlap and the
disagreement rates among the ensemble predictions are highest. By contrast,
AdaBoost tends to give more weight to those instances whose class label has been
modified. Focusing on these noisy instances is misleading and eventually impairs
the generalization capacity of AdaBoost. To illustrate this observation, the
experiment was repeated injecting class-label noise in the training data.
Specifically, 30\% of the training examples were selected at random and their
class labels flipped, as in the noisy completely at random model described in
\cite{frenay_2014_classification}. The results of these experiments are shown in
\figurename~\ref{fig_wn}. Instances whose class label has been switched are
marked with a red cross in these plots. For $a = b = 1.0$, no correlation is
observed between the ranks of the weights given by vote-boosting and AdaBoost.
However, instances whose class label has been switched, which are distributed
uniformly in the vertical direction, tend to appear on the right-hand side of
the plots. This means that they receive special emphasis in AdaBoost but not in
vote-boosting. Increasing the value of $a = b $ pushes the unperturbed instances
towards the diagonal but not the perturbed ones. However, the correlation
between the ranks is less prominent than in the noiseless case, because of the
interference of the noisy instances. 

\begin{figure*}
\centering
\subfloat[]{\includegraphics[scale=0.35]{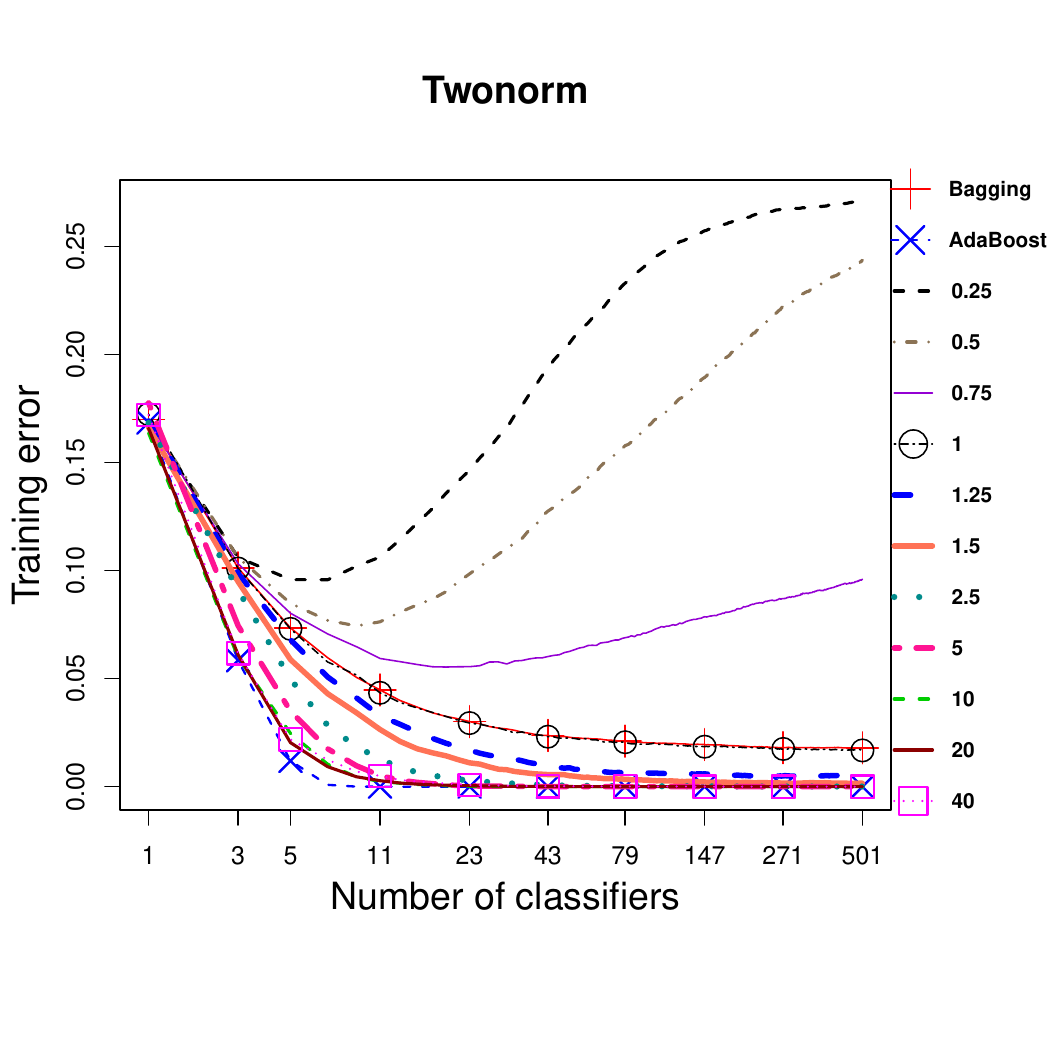}
	}
\hfil
\subfloat[]{\includegraphics[scale=0.35]{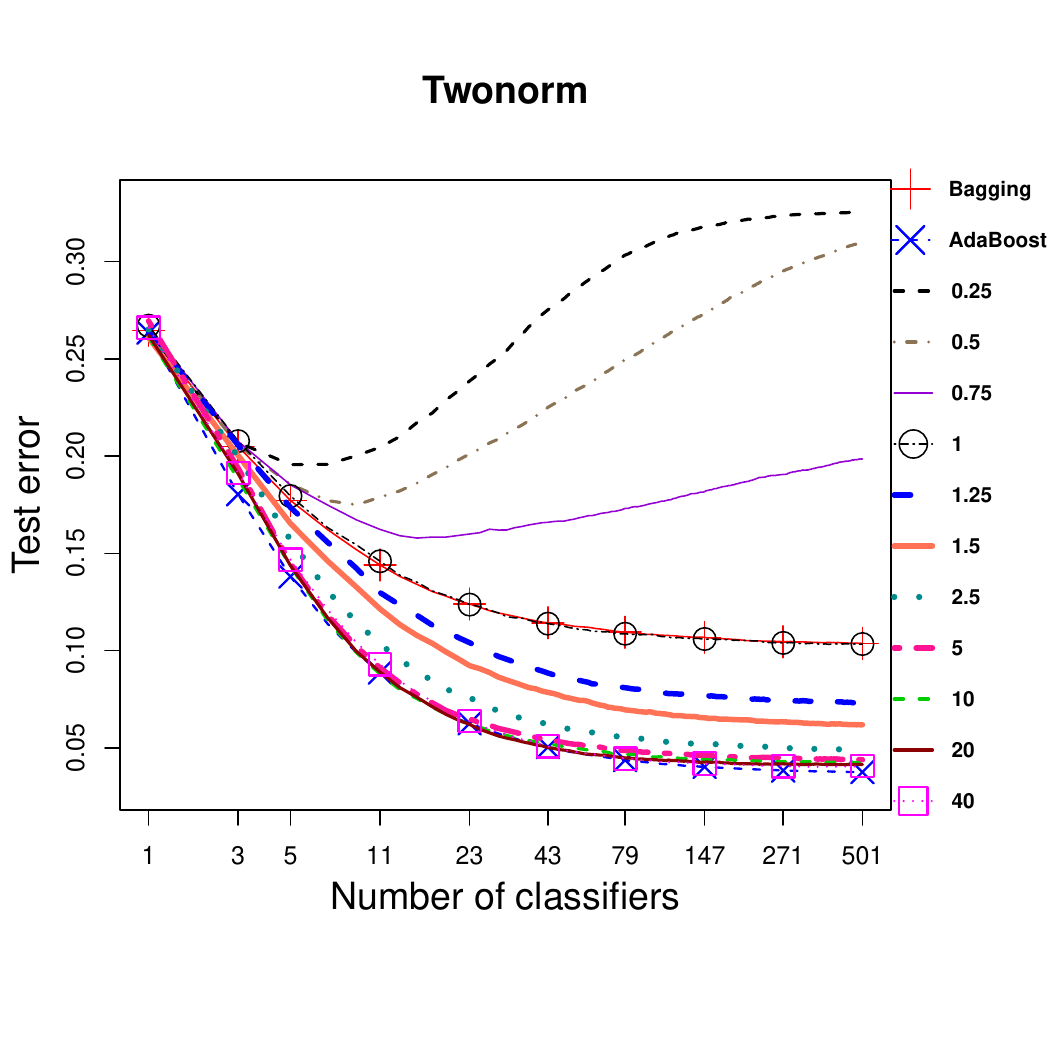}
	}
\caption{Error rate as a function of number of classifiers in {\it Twonorm} for
bagging, AdaBoost, and vote-boosting ensembles of pruned CART trees: (a)
training error (b) test error.} \label{interpol-twonorm} 
\end{figure*}
		
\begin{figure*}
\centering
\subfloat[]{\includegraphics[scale=0.35]{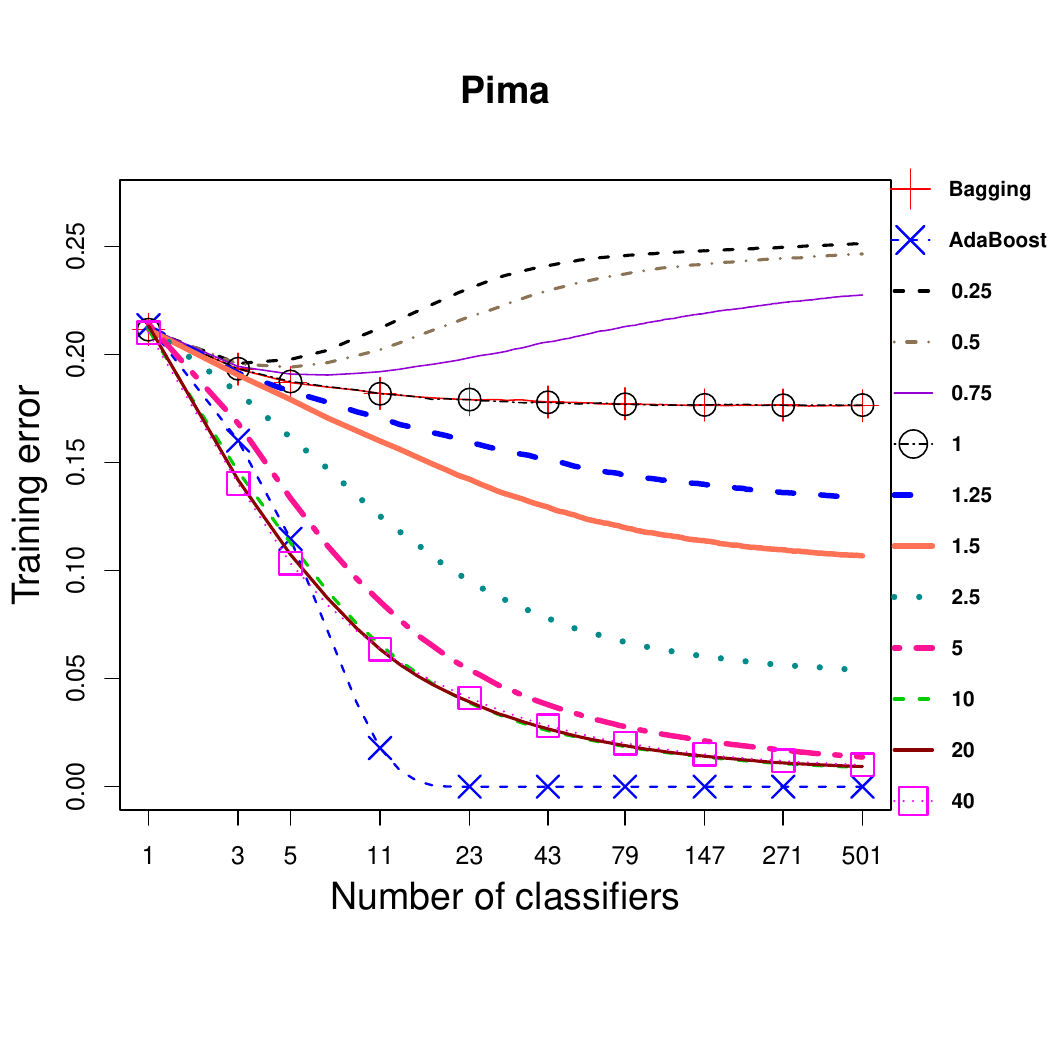}
	\label{fig_first_case}}
\hfil
\subfloat[]{\includegraphics[scale=0.35]{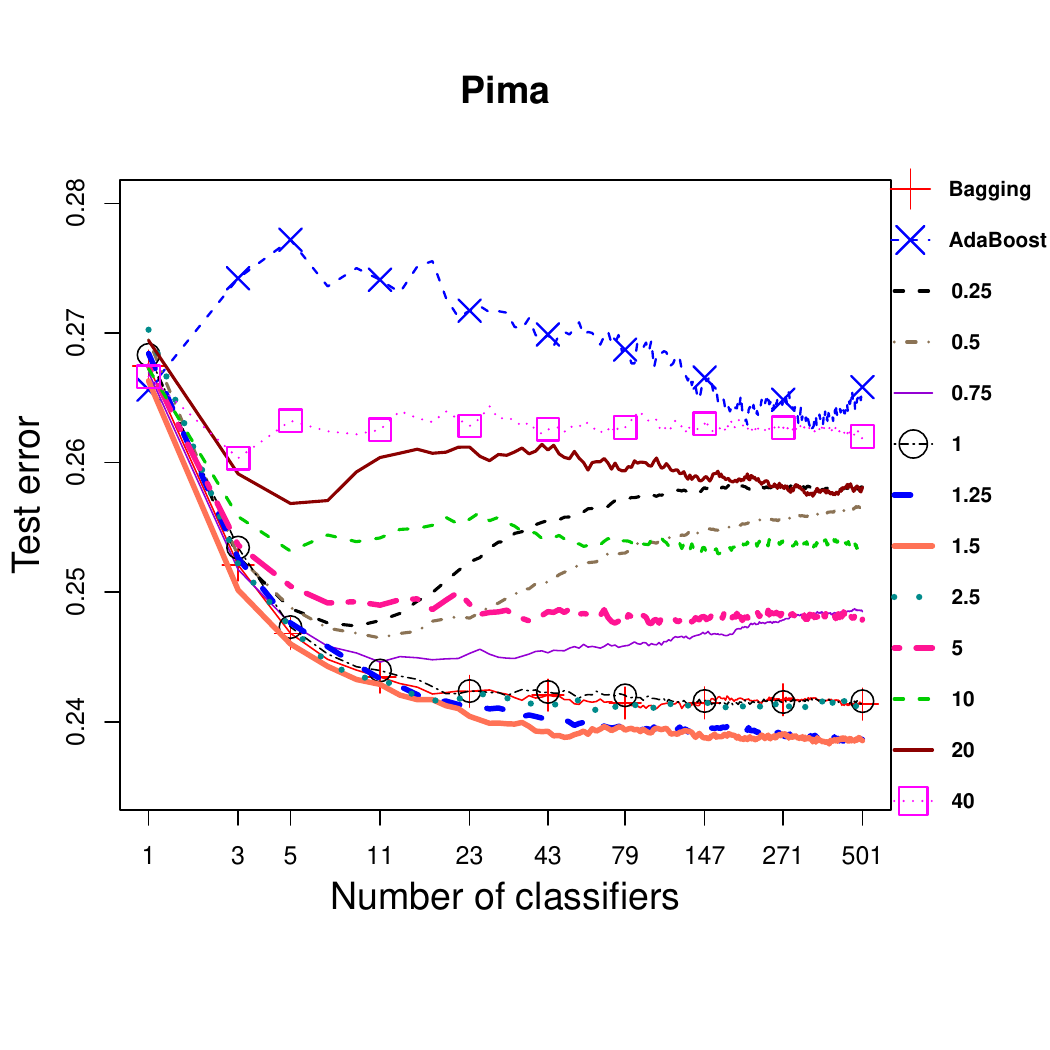}
	\label{fig_second_case}}
      \caption{Error rate as a function of the number of classifiers in {\it Pima}
for bagging, AdaBoost, and vote-boosting ensembles of pruned CART trees: (a)
training error (b) test error.} \label{interpol-pima} 
\end{figure*}
				
A second batch of experiments was carried out to analyze the behavior of
vote-boosting composed of pruned CART trees
as a function of the strength of the emphasis that is applied to
uncertain instances. Using the symmetric beta distribution as emphasis function,
we analyze how the learning curves, which trace the dependence of the error as a
function of the size of the ensemble, depend on the value of the shape
parameter. The values explored are $a$=$b \in \{0.25, 0.5, 0.75, 1.0, 1.25, 1.5,
2.5, 5, 10, 20, 40 \}$. The experiments were made on the classification tasks {\it
Twonorm} and {\it Pima}. These tasks have been chosen because of the different
prediction accuracies of bagging and AdaBoost on those datasets. In {\it
Twonorm}, AdaBoost significantly outperforms bagging. By contrast in {\it Pima},
which is a very noisy task, bagging is more accurate than AdaBoost.
		
\figurename~\ref{interpol-twonorm} and \ref{interpol-pima} display the learning
curves for bagging, AdaBoost, and vote-boosting using different values of the
shape parameter in the classification tasks {\it Twonorm} and {\it Pima},
respectively. The plots on the left-hand side show the results for the training
error. The plots on the right-hand side correspond to the test error curves.
When $a = b = 1.0$, all instances are given equal weights. In this case, if
weighted resampling is used, vote-boosting is equivalent to bagging. 
This is apparent from \figurename~\ref{interpol-twonorm} and \ref{interpol-pima}:
the error curves of both methods are very close to each other.
When values $ a = b < 1.0$ are used, emphasis is made on instances on which
most predictors agree. If regularized classifiers, such as pruned CART
trees, are used as base learners, this type of emphasis is in general not
effective. Nonetheless, as will be illustrated by the results presented in
Section \ref{sec:performance}, focusing on these types of instances can lead to
improvements in the generalization capacity when the data are very noisy and the
ensemble is composed of flexible classifiers that overfit (e.g. unpruned
CART or random trees).

Values of $ a = b > 1.0$ correspond to emphasizing instances in which the
ensemble prediction is uncertain. For {\it Twonorm}, the learning curves of
vote-boosting using $a = b = 40.0$  and AdaBoost are quite similar. 
In this problem,
the classification errors are mainly due to the overlap
between the distributions of the two classes. Therefore, the incorrectly
classified instances are close to the decision boundary, where the ensemble
predictions are also more uncertain. Using a beta distribution sharply peaked 
around $\pi = 0.5$ (see
\figurename~\ref{sym-beta}) gives more weight to these uncertain instances. In
consequence, the resulting emphasis is similar to AdaBoost's. In {\it Pima},
which is a noisy problem, the learning curves of vote-boosting with large $a =
b$ and AdaBoost are different. Still, the closest performance to AdaBoost is
vote-boosting with large $a=b$. Finally, we observe that the optimal value for
the shape parameter of the beta distribution in vote-boosting is problem
dependent: Values of $a = b \approx 1.25-1.5$ perform well in {\it Pima}. The
best performance in {\it Twonorm} requires using a large values of $a = b
\approx 40$. For each problem, the optimal value can be determined using
cross-validation.

\subsection{Vote-boosting with different base learners}
 
In this section, we present the results of a comparison
of vote-boosting ensembles composed of different base learners.
These are, in order of increasing complexity, 
decision stumps, pruned and unpruned CART trees, and unpruned random trees. 
Ensembles composed of $501$ classifiers are built. This fairly large ensemble
size is needed in some problems to achieve convergence to the asymptotic error
level \cite{hernandezLobato++_2011_inference}. 
Weighted resampling with replacement is used to generate the bootstrap 
samples on which the individual classifiers are trained.
In vote-boosting, the symmetric beta distribution is used for emphasis. 
The shape parameter is determined as the value among those 
in $a$=$b \in$ \{0.25, 0.5, 0.75, 1.0, 1.25,
1.5, 2.5, 5, 10, 20, 40\} that minimizes the 10-fold cross-validation 
in the training set.
When uniform emphasis in all the training instances ($a=b=1.0$) is used,
the results are equivalent to bagging or, if random trees are used 
as base learners, to random forest.
For values of the shape parameter above $1.0$, the symmetric beta distribution has
a single mode at $\pi = 0.5$, which implies that more emphasis is made on
training instances on which the degree of disagreement among the different
classifiers is large. 
By contrast, when the shape parameter is smaller
than $1$ the focus is on training instances on which most classifiers agree.

For the empirical evaluation carried out
in this an the following section, 
different binary classification tasks from
the UCI repository \cite{Bache+Lichman:2013} and other sources
\cite{breiman_1998_arcing} are considered. The characteristics of the datasets 
used in this study are summarized in Table~\ref{tab:datasets}. For each
dataset, the table displays the total number of labeled instances available,
the number of those instances used for training and for testing, 
and the number of attributes.
The test error rates reported are averages, followed by the
corresponding standard deviations after the $\pm$ sign, over $100$
realizations of the training and test sets. For classification
problems in which only a finite collection of labeled instances is available, $2/3$ of
the data are selected at random for training and the remaining $1/3$ for
testing. For synthetic problems (namely, {\it Ringnorm}, {\it Twonorm}, and 
{\it Threenorm}), instances are generated independently at
random: $300$ instances are used for training and $2000$ for testing. In all
cases, stratified sampling is used to ensure that the class distributions in the
test and training sets are similar.
 
\begin{table}[t!] \caption{Characteristics of the classification problems
analyzed and training / test partitions.}
	\label{tab:datasets}
\begin{center}
	\begin{tabular}{lcccc}
		\hline
		\multicolumn{1}{l}{\bf Dataset}&
		\multicolumn{1}{c}{\bf Instances}&
		\multicolumn{1}{c}{\bf Training}&
		\multicolumn{1}{c}{\bf Test}&
					{\bf Number of}
		\\
		\multicolumn{1}{l}{}&
		\multicolumn{1}{c}{}&
		\multicolumn{1}{c}{}&
		\multicolumn{1}{c}{}&
					{\bf attributes}
		\\
		\hline
					Adult       & 32561 & 21707 & 10854     & 15   \\
					Australian  & 690   & 460   & 230       & 14   \\
					Breast W.   & 699   & 466   & 233       & 9    \\
					Blood       & 748   & 499   & 249       & 5    \\
					Boston      & 506   & 337   & 169       & 14   \\
					Chess       & 3196  & 2131  & 1065      & 36   \\
					German      & 1000  & 667   & 333       & 20   \\
					Heart       & 270   & 178   & 92        & 13   \\
					Hepatitis   & 155   & 104   & 51        & 19   \\
					Horse-colic & 368   & 246   & 122       & 21   \\
					Ionosphere  & 351   & 234   & 117       & 34   \\
					Liver       & 345   & 230   & 115       & 6    \\
				  Magic       & 19020 & 12680 & 6340      & 11   \\
					Musk        &6598   & 4399  & 2199      & 168  \\
					Ozone       & 2536  & 1691  & 845       & 74   \\
					Parkinsons  & 197   & 132   & 65        & 24   \\
					Pima        & 768   & 512   & 256       & 8    \\
					Ringnorm    & 2300  & 300   & 2000      & 20   \\
					Sonar       & 208   & 139   & 69        & 60   \\
					Spambase    & 4601  & 3068  & 1533      & 58   \\
					Threenorm   & 2300  & 300   & 2000      & 20   \\
					Tic-tac-toe & 958   & 639   & 319       & 9    \\
					Twonorm     & 2300  & 300   & 2000      & 20   \\
		\hline
	\end{tabular}
\end{center}
\end{table}

In Table~\ref{tab:results1_base}, 
the average test errors over the $100$ realizations are shown for all base learners.
For each dataset, the lowest average generalization error is highlighted in 
boldface and the second best is underlined. 
If the differences between the two lowest 
test errors is statistically significant at a significance level $\alpha =0.05$
the lowest error value is marked with an asterisk (*). 
A resampled paired t-test is used for synthetic problems.
When random train/test partitions of the same dataset
are employed a corrected resampled paired t-test 
\cite{nadeau+bengio_2003_inference,bouckaert+frank_2004_evaluating}
is used instead. 

\begin{figure*}[!t]
		\centering
		\includegraphics[scale=0.475]{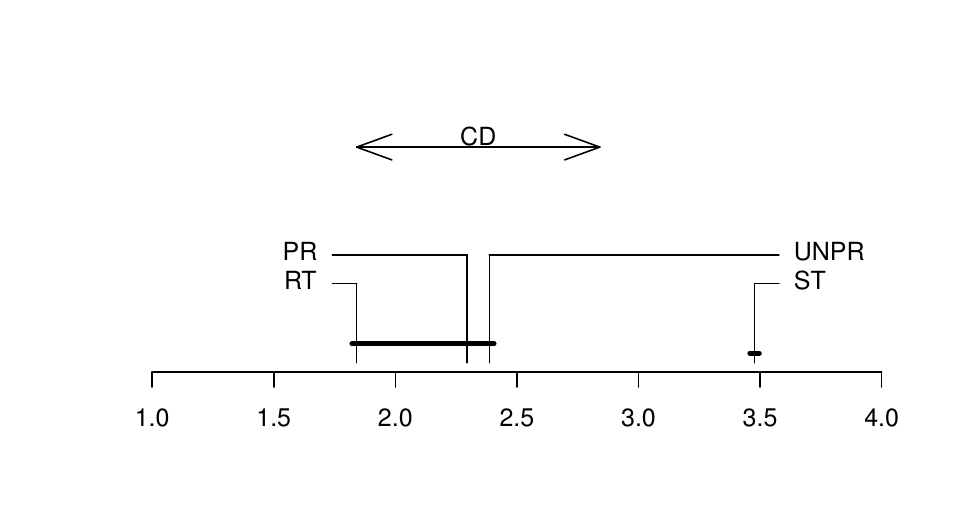}
		\caption{Comparison of the average ranks of decision
			stumps (ST), pruned CART trees (PR), unpruned CART trees (UNPR), and random trees (RT) using a Nemenyi test. Horizontal lines connect methods whose average ranks are not significantly different (p-value $<$ 0.05).} \label{demsar_base}
	\end{figure*}

To provide an overall comparison of the accuracies of vote-boosting using the
four tested base learners, we apply the framework proposed in
\cite{demsar_2006_statistical}. To this end, the average rank of the classifier
average errors is computed for the $23$ classification problems investigated.
The rank of a classifier in a specific classification problem is determined by
ordering the different methods according to their test errors. A lower rank
corresponds to a smaller test error and, therefore, better accuracy. The average
ranks of vote-boosting using the four different base learners are displayed in
\figurename~\ref{demsar_base}. In this diagram, the differences of average ranks
between methods that are connected by a horizontal solid line are not
statistically significant according to a Nemenyi test (p-value $<$ 0.05). 

From the results presented in Table~\ref{tab:results1_base} 
and \figurename~\ref{demsar_base}, one concludes that 
vote-boosting composed of random trees has the 
best overall accuracy:
except in {\it Boston}, {\it
Horse-colic}, and {\it Parkinsons}, 
these types of ensembles achieve the lowest
or second lowest average test errors. 
Notwithstanding, according to the Nemenyi test, 
the average rank differences  
are statistically significant only with respect to the use of stumps
(see \figurename~\ref{demsar_base}). 

The values of the shape parameters of the symmetric beta distribution ($a=b$)
selected by cross-validation are shown in Table~\ref{tab:results2_base}.
The figures reported are medians over the $100$ realizations
of the training and test data. 
An asterisk is shown in the Table when ensembles built using
the different values have the same within-train cross-validation error. 
For decision stumps, this occurs in {\it Australian} and {\it Musk} 
because the individual stumps are the same 
irrespective of the type of emphasis employed.
In {\it Musk}, when more complex base classifiers are used, the  
ensembles built with the different values of $a=b$ all achieve zero error.
In most cases, the value of the shape parameter of the beta distribution
($a=b$) decreases as the complexity of the base classifier increases. 
The reason of this is that more emphasis is 
needed in order to \emph{boost} more stable base classifiers. 
Thus, the largest values of $a=b$ are selected for decision stumps.
Hence, for these types of base learners the focus is 
on uncertain instances, on which most ensemble classifiers disagree. 
The lowest values of $a=b$ correspond to random trees. In 
this case, the emphasis on uncertain instances is reduced.
In fact, for {\it Blood}, {\it Heart}, {\it Musk}, and {\it Pima},
the focus should be on instances on which the ensemble 
classifiers agree.

In summary, vote-boosting ensembles composed of random trees provide the best
overall accuracy in the problems investigated. Since this type of ensemble can
also be built efficiently, they will be used for further evaluation in the
following set of experiments.

\begin{table}
	
	\caption{Test error rates of vote-boosting ensembles composed of decision stumps, pruned CART trees, unpruned CART trees and random trees.}
	\label{tab:results1_base}
	\centering{%
			
	\begin{tabular}
			{lr@{$\pm$}lr@{$\pm$}lr@{$\pm$}lr@{$\pm$}lr@{$\pm$}l}
			\hline
			
			Dataset  &
			\multicolumn{2}{c}{Stump} &
			\multicolumn{2}{c}{Pruned} &
			\multicolumn{2}{l}{Unpruned} &
			\multicolumn{2}{l}{Random tree}  \\ 
					\hline
					Adult  & 20.9 & 2.9& 13.6 & 0.3& {\bf 13.5} & {\bf 6.2}& \underline{13.6} & \underline{0.2} \\
					Australian
					& 14.6 & 2.1 & 13.6 & 2.0& \underline{13.5} & \underline{2.1}& {\bf 13.3} & {\bf 2.1}\\
					Breast W.
					& 3.9 & 1.0& \underline{3.8} & \underline{1.1} & 4.0 & 1.2& {\bf 3.1} & {\bf 1.1}\\
					Blood
					& 22.7 & 1.5 & 22.2 & 1.7& {\bf 21.8} & {\bf 1.8}& \underline{21.9} & \underline{1.7}\\
					Boston
					& 15.2 & 2.3& {\bf 12.7} & {\bf 2.5}& \underline{13.0} & \underline{2.6} & 13.1 & 2.3\\
					Chess&  17.7 & 14.1& \underline{ 0.6} & \underline{ 0.6}& 0.7 & 0.6& {\bf 0.6} & {\bf 0.3}\\ 
					German
					& 25.4 & 1.6& \underline{ 24.1} & \underline{ 2.1}& {\bf 24.1} & {\bf 2.0}& 24.3 & 1.7\\
					Heart
					&  {\bf 16.4} & {\bf 3.5} & 17.8 & 3.6 & 18.4 & 3.9& \underline{16.8} & \underline{3.5}\\
					Hepatitis  & 16.4 & 4.3& \underline{16.2} & \underline{4.8} & 16.9 & 4.9& {\bf 14.0} & {\bf 3.9}\\
					Horse-Colic
					& {\bf 13.9} & {\bf 2.7}& 14.1 & 2.9& \underline{ 13.9} & \underline{ 2.7} & 15.0 & 2.5\\
					Ionosphere&   8.1 & 1.7& \underline{6.7} & \underline{2.0}& 6.7 & 1.8& {\bf 6.6} & {\bf 1.8}\\
					Liver& {\bf 27.5} & {\bf 3.9}& 28.4 & 3.6 & 28.6 & 3.9& \underline{28.4} & \underline{3.7}\\
					Magic&  15.4 & 0.3& \underline{12.7} & \underline{0.3} & 12.9 & 0.3& {\bf 11.6} & {\bf 0.2}*\\
					Musk&  {\bf 0.0} & {\bf 0.0}& {\bf 0.0} & {\bf 0.0}& {\bf 0.0} & {\bf 0.0}& {\bf 0.0} & {\bf 0.0} \\
					Ozone& 6.3 & 0.0& {\bf 5.8} & {\bf 0.4}& \underline{ 5.8} & \underline{0.4}& 6.0 & 0.2\\
					Parkinsons & 11.5 & 4.3& {\bf 7.4} & {\bf 3.6}& \underline{7.6} & \underline{3.2} & 9.3 & 3.6\\
					Pima
					& 24.1 & 2.0 & 24.2 & 2.2& \underline{23.8} & \underline{2.3}& {\bf 23.4} & {\bf 1.8}\\
					Ringnorm & 9.3 & 0.9& \underline{4.5} & \underline{0.8} & 4.6 & 0.8& {\bf 4.4} & {\bf 0.8}\\
					Sonar& 18.9 & 4.6 & 16.0 & 4.4& \underline{15.8} & \underline{4.8}& {\bf 15.5} & {\bf 4.5}\\
					Spambase& 6.1 & 0.6& 4.5 & 0.3& \underline{4.5} & \underline{0.3}& {\bf 4.1} & {\bf 0.5} \\
					Threenorm& 20.1 & 1.0 & 17.1 & 1.1& \underline{17.0} & \underline{1.0}& {\bf 16.2} & {\bf 1.0}*\\
					Tictactoe & 22.0 & 2.9& {\bf 0.7} & {\bf 0.6}& \underline{ 0.7} & \underline{ 0.6}& 1.1 & 0.7\\
					Twonorm& 4.6 & 0.7& 4.2 & 0.6& \underline{4.2} & \underline{0.6}& {\bf 3.6} & {\bf 0.5}*\\
					\hline
					
		\end{tabular}}
	\end{table}
	
	\begin{table}[ht!]
    \caption{Median of the beta distribution parameter ensembles composed of decision stumps, pruned CART trees, unpruned CART trees, and random trees. 
An asterisk is shown when all values of the beta parameter yield the same
cross-validation error}
			\label{tab:results2_base}
			\centering{%
		\begin{tabular}{l  r  r  r  r}
			\hline
			Dataset  &
			{Stump}  & {Pruned}&
			{Unpruned}&
			{Random tree}\\
			\hline
			Adult&40.0&40.0&20.0&1.0\\
			Australian&* &    5.0    &    2.5  &    1.5  \\
			Breast W. &10.0 &  5.0  &  10.0 &  1.0  \\
			Blood &40.0    &     1.25  &     0.75  &     0.5 \\
			Boston & 40.0 &  20.0 &  10.0 &  2.0  \\
			Chess& 20.0  &   20.0  &   20.0  &  10.0  \\
			German &40.0   &    5.0 &    2.5  &    2.5    \\ 
			Heart &10.0  &   2.5 &   1.5 &   0.5 \\
			Hepatitis& 10.0  &   10.0  &   5.0   &   2.5 \\
			Horse-Colic &20.0  &   1.5 &   1.5 &   1.5 \\
			Ionosphere&40.0 &  5.0  &  10.0 &  2.5  \\
			Liver &40.0  &   5.0   &   5.0   &   1.5 \\
			Magic&40.0 &  40.0 &  40.0 &  20.0 \\
			Musk&*& * & * &    0.7\\
			Ozone&40.0  &   10.0  &   5.0 &   1.5 \\ 
			Parkinsons&40.0 &  15.0 &  10.0 &  5.0 \\
			Pima& 15.0  &   1.5 &   1.0   &   0.5 \\
			Ringnorm&40.0 &  20.0 &  20.0 &  20.0 \\
			Sonar&40.0 &  20.0 &  20.0 &  10.0 \\
			Spambase& 40.0 &  20.0 &  20.0 &  20.0   \\
			Threenorm&40.0 &  10.0 &  10.0 &  5.0  \\
			Tictactoe&40.0 &  20.0 &  20.0 &  10.0 \\
			Twonorm&20.0 &  20.0 &  20.0&  5.0  \\
			\hline
		\end{tabular}}
	\end{table}
	
\subsection{Comparison of vote-boosting with other ensemble methods} \label{sec:performance}
In this section, we carry out an comparison of 
vote-boosting ensembles and other related methods: bagging, AdaBoost and random
forest. The classification problems considered and experimental protocol
followed are the same as in the previous section. In each type of ensemble, the
base classifier that performs best is used: pruned CART trees in AdaBoost,
unpruned CART trees in bagging, and random trees in vote-boosting and random
forest. Note that unpruned CART or random trees cannot be used in combination
with AdaBoost because they generally achieve zero training error, which gives
rise to singularities in the weight updates. As illustrated in the previous
section, similar results are obtained with vote-boosting ensembles composed of
unpruned or pruned CART trees.  The adabag \cite{alfaro++_2013_adabag}, ipred
\cite{peters+hothorn_2009_ipred} and ranfomForest
\cite{liaw+wiener_2002_classification} packages in R have been used for
AdaBoost, bagging, and random forest implementations, respectively. As in the
previous subsection, ensembles of $501$ classifiers are used to guarantee
convergence to the asymptotic error level. In both vote-boosting and AdaBoost,
weighted resampling with replacement is used to take into account the different
emphasis on the training instances. In AdaBoost, reweighting was considered as
an alternative. However, as reported in the literature, similar or slightly
better results are obtained when resampling instead of reweighting is used
\cite{dietterich_2000_experimental,seiffert++_2008_resampling}. 

The experiments are carried out on the original problem, and also with 10\%,
20\% and 30\% class-label noise. Class-label noise is injected into the training
set by randomly switching the class label of a random subset (10\%, 20\% and
30\%) of the training instances. This type of label noise is know as completely
at random noise (NCAR) \cite{frenay_2014_classification}. An interesting
observation is that, as the noise level increases, the values of the shape
parameter that are selected tend to decrease. 

	\begin{table}
		
		\caption{Test error rates for bagging (unpruned CART trees), boosting (pruned CART trees), random forest,
		and vote-boosting (random trees).}
		\label{tab:results1}
		\centering \resizebox{\linewidth}{!}{%
			
			\begin{tabular}
				{@{}l@{$\;$}|r|r@{$\pm$}l@{$\;$}|r@{$\pm$}l@{$\;$}|r@{$\pm$}l@{$\;$}|r@{$\pm$}l@{$\;$}|p{1.1cm}c@{}}
				\hline
				& Noise   \\
				Dataset  & (in \%) &
				\multicolumn{2}{c|}{Bagging} &
				\multicolumn{2}{c|}{AdaBoost} &
				\multicolumn{2}{c|}{Random forest} &
				\multicolumn{2}{c|}{Vote-boosting} &
				\multicolumn{2}{c}{a=b (median)}  \\ 
		\hline
		Adult
		& 0& 14.6&0.2& 13.8 & 0.2& \underline{13.6} & \underline{ 0.2}& {\bf 13.6} & {\bf 0.2}&1.0\\
		& 10&15.5&0.3&14.3& 0.3 &\underline{14.1} &\underline{0.3}& {\bf 14.0} & {\bf 0.3}&0.75\\
		& 20&17.0&0.3&\underline{14.8}&\underline{0.2}& 14.8 & 0.3& {\bf 14.3} & {\bf 0.2}&0.25\\
		& 30&20.3&0.3&\underline{ 15.1}&\underline{ 0.3} & 17.8 & 0.3& {\bf 15.1} & {\bf 0.2}&0.25\\
		\hline
		Australian
		& 0& 13.3 & 2.1 & 13.7 & 1.9& {\bf 13.0} & {\bf 2.0}& \underline{13.3} & \underline{2.1}& 1.5 \\
		& 10& 14.9 & 2.3 & 17.6 & 2.4& {\bf 13.6} & {\bf 2.0}& \underline{ 13.6} & \underline{ 2.1}& 0.75 \\
		& 20 & 18.2 & 2.9 & 24.3 & 3.0& \underline{15.8} & \underline{2.2}& {\bf 14.3} & {\bf 2.4}*& 0.25 \\
		& 30 & 24.6 & 3.8 & 32.0 & 3.5& \underline{21.6} & \underline{3.0}& {\bf 18.2} & {\bf 3.3}*& 0.25 \\
		\hline
		Breast W.
		& 0 & 4.6 & 1.2 & 3.6 & 1.0& {\bf 3.0} & {\bf 1.0}& \underline{3.1} & \underline{1.1} &  1.0 \\
		& 10 & 6.3 & 1.6 & 6.9 & 1.8& \underline{4.1} & \underline{1.2}& {\bf 3.5} & {\bf 1.2}* &  0.5 \\
		& 20 & 9.6 & 2.5 & 12.0 & 2.6& \underline{6.5} & \underline{1.9}& {\bf 4.1} & {\bf 1.4}* &  0.25 \\
		& 30 & 17.7 & 3.0 & 20.9 & 3.4& \underline{13.0} & \underline{2.7}& {\bf 6.8} & {\bf 2.6}* &  0.25 \\
		\hline
		Blood
		& 0 & 26.3 & 1.9 & 25.4 & 2.1& \underline{24.5} & \underline{2.2}& {\bf 21.9} & {\bf 1.7}* &  0.5 \\
		& 10 & 28.5 & 2.2 & 27.2 & 2.5& \underline{26.6} & \underline{2.4}& {\bf 22.6} & {\bf 2.0}* &  0.25 \\
		& 20 & 31.6 & 2.7 & 30.2 & 2.6& \underline{29.7} & \underline{2.6}& {\bf 23.8} & {\bf 2.5}* &  0.25 \\
		& 30 & 35.8 & 3.6 & 34.6 & 3.8& \underline{34.2} & \underline{3.6}& {\bf 29.0} & {\bf 4.5}* &  0.25 \\
		\hline
		Boston
		& 0 & 14.2 & 2.4& {\bf 12.7} & {\bf 2.4}& \underline{13.0} & \underline{2.4} & 13.1 & 2.3 &  2.0 \\
		& 10 & 16.2 & 2.8 & 16.9 & 2.5& {\bf 14.5} & {\bf 2.6}& \underline{14.9} & \underline{2.5} &  0.5 \\
		& 20 & 19.1 & 3.3 & 22.2 & 3.6& \underline{17.5} & \underline{3.2}& {\bf 16.2} & {\bf 2.8} &  0.5 \\
		& 30 & 26.4 & 4.2 & 31.4 & 4.0& \underline{24.6} & \underline{3.8}& {\bf 21.3} & {\bf 4.4}* &  0.25 \\
		\hline
		Chess
		& 0& 0.6 & 0.3& {\bf 0.5} & {\bf 0.3} & 1.6 & 0.4& \underline{0.6} & \underline{0.3}& 10.0 \\
		& 10 & 5.1 & 0.7& 2.3 & 0.5& \underline{ 2.1} & \underline{ 0.5}& {\bf 2.1} & {\bf 0.5}& 1.25 \\
		& 20 & 11.9 & 1.2& {\bf 3.6} & {\bf 0.7} & 4.3 & 0.9& \underline{4.0} & \underline{0.9}& 0.75 \\
		& 30 & 22.4 & 1.5& {\bf 5.5} & {\bf 0.9}* & 10.2 & 1.3& \underline{7.9} & \underline{1.5}& 0.25 \\
		\hline
		German
		& 0 & 24.5 & 2.0 & 24.8 & 2.1& {\bf 24.0} & {\bf 1.8}& \underline{24.3} & \underline{1.7} &  2.5 \\
		& 10& 26.2 & 2.1 & 27.9 & 2.1& \underline{ 25.3} & \underline{ 1.8}& {\bf 25.3} & {\bf 1.9} &  1.0 \\
		& 20 & 28.8 & 2.4 & 32.2 & 2.5& {\bf 27.3} & {\bf 2.2}& \underline{27.4} & \underline{2.3} &  0.75 \\
		& 30 & 32.3 & 3.0 & 37.4 & 2.8& \underline{31.0} & \underline{3.1}& {\bf 29.9} & {\bf 3.2} &  0.5 \\
		\hline
		Heart
		
		& 0 & 19.5 & 3.7 & 20.2 & 3.4& \underline{17.2} & \underline{3.0}& {\bf 16.8} & {\bf 3.5} &  0.5 \\
		& 10 & 22.7 & 4.2 & 24.9 & 4.5& \underline{19.3} & \underline{3.7}& {\bf 18.7} & {\bf 4.0} &  0.5 \\
		& 20 & 26.0 & 5.0 & 30.6 & 5.1& \underline{22.7} & \underline{4.2}& {\bf 20.8} & {\bf 4.0} &  0.375\\
		& 30 & 32.1 & 5.4 & 36.3 & 6.0& \underline{28.5} & \underline{5.8}& {\bf 25.8} & {\bf 6.6} &  0.25 \\
		\hline
		Hepatitis
		
		& 0 & 15.8 & 4.7 & 15.8 & 4.0& {\bf 13.6} & {\bf 3.5}& \underline{14.0} & \underline{3.9} &  2.5 \\
		& 10 & 17.1 & 4.7 & 20.0 & 5.0& {\bf 14.2} & {\bf 3.9}& \underline{15.3} & \underline{3.9} &  1.0 \\
		& 20 & 21.0 & 5.8 & 26.0 & 6.9& \underline{17.6} & \underline{4.8}& {\bf 17.3} & {\bf 4.5} &  0.5 \\
		& 30 & 27.8 & 7.8 & 33.6 & 7.2& \underline{23.8} & \underline{7.3}& {\bf 21.9} & {\bf 7.0} &  0.5 \\
		\hline
		Horse-colic
		& 0& {\bf 14.9} & {\bf 2.8} & 15.4 & 2.6& \underline{15.0} & \underline{2.5}& 15.0 & 2.5 &  1.5 \\
		& 10 & 17.8 & 3.5 & 21.2 & 3.5& {\bf 17.3} & {\bf 2.9}& \underline{17.4} & \underline{3.0} &  1.0 \\
		& 20 & 23.4 & 4.2 & 27.6 & 4.7& \underline{21.3} & \underline{3.6}& {\bf 20.9} & {\bf 3.7} &  0.625 \\
		& 30 & 30.0 & 5.6 & 34.9 & 5.1& \underline{28.4} & \underline{4.6}& {\bf 28.2} & {\bf 5.3} &  0.75 \\
		\hline
				\end{tabular}}
				\end{table}
	
	\begin{table}
		
		\caption{Test error rates for bagging (unpruned CART trees), boosting (pruned CART trees), random forest, and vote-boosting (random trees).}
		\label{tab:results2}
		\centering \resizebox{\linewidth}{!}{%
			
			\begin{tabular}
				{@{}l@{$\;$}|r|r@{$\pm$}l@{$\;$}|r@{$\pm$}l@{$\;$}|r@{$\pm$}l@{$\;$}|r@{$\pm$}l@{$\;$}|p{1.1cm}c@{}}
				\hline
				& Noise   \\
				Dataset  & (in \%) &
				\multicolumn{2}{c|}{Bagging} &
				\multicolumn{2}{c|}{AdaBoost} &
				\multicolumn{2}{c|}{Random forest} &
				\multicolumn{2}{c|}{Vote-boosting} &
				\multicolumn{2}{c}{a=b (median)}  \\ 
				\hline
					Ionosphere
					& 0& 8.0 & 2.2& {\bf 6.6} & {\bf 1.8}&  6.6 &  1.6& \underline{ 6.6} & \underline{ 1.8} &  2.5 \\
					& 10& 9.5 & 2.7 & 10.5 & 2.6& \underline{ 7.9} & \underline{ 2.2}& {\bf 7.9} & {\bf 2.2} &  0.75 \\
					& 20 & 13.1 & 3.1 & 17.0 & 3.5& \underline{10.9} & \underline{2.9}& {\bf 9.9} & {\bf 3.1} &  0.5 \\
					& 30 & 19.7 & 4.5 & 26.4 & 4.5& \underline{17.8} & \underline{4.6}& {\bf 15.7} & {\bf 5.1} &  0.25 \\
					\hline
					Liver
					& 0 & 29.5 & 3.9 & 30.5 & 3.9& {\bf 27.7} & {\bf 3.8}& \underline{28.4} & \underline{3.7} &  1.5 \\
					& 10 & 31.7 & 4.1 & 33.8 & 4.5& {\bf 31.0} & {\bf 4.0}& \underline{31.4} & \underline{3.8} &  1.0 \\
					& 20 & 36.1 & 4.3 & 37.8 & 4.5& {\bf 35.1} & {\bf 4.4}& \underline{35.7} & \underline{4.4} &  1.0 \\
					& 30 & 39.9 & 4.7 & 41.5 & 4.7& {\bf 39.6} & {\bf 4.5}& \underline{39.7} & \underline{4.9} &  0.875\\
					\hline
					Magic
					& 0& 12.3& 0.3& 12.9& 0.3 &\underline{12.0} & \underline{0.3}& {\bf 11.6} & {\bf 0.2}*&20.0\\
					& 10&12.9&0.3&13.9& 0.3& {\bf 12.4} & {\bf 0.2}& \underline{12.5} & \underline{0.2}&1.25\\
					& 20& 14.2&0.4& 14.4 &0.3&\underline{13.6} & \underline{0.4}& {\bf 13.3} & {\bf 0.4}*&0.5\\
					& 30& 17.4&0.5&\underline{15.0}& \underline{0.5}&16.7 & 0.4& {\bf 14.6} & {\bf 0.4}*&0.25\\
					\hline
					Musk
					&0&{\bf 0.0}&{\bf 0.0}& {\bf 0.0}& {\bf 0.0}& {\bf 0.0}& {\bf0.0}&{\bf0.0}&{\bf 0.0}& 0.7\\
					& 10& 1.4 & 0.3 & \underline{0.6} & \underline{0.2} & 1.6 & 0.3 & {\bf 0.1} & {\bf 0.1}*&0.25\\
					& 20& 4.5&0.4&\underline{ 0.9} & \underline{ 0.6}&5.7 & 0.4& {\bf 0.8} & {\bf 0.3}&0.25\\
					& 30& 12.0 & 1.0 &{\bf 2.5} & {\bf 1.2}*& 14.4 & 1.2&  \underline{5.1} &  \underline{0.7}&0.25\\
					\hline
					Ozone
					& 0& 6.0 & 0.3 & {\bf 5.7} & {\bf 0.4}*& 6.0 & 0.2& \underline{6.0} & \underline{0.2} &  1.5 \\
					& 10& 6.2 & 0.3 & 6.4 & 0.5& \underline{ 6.0} & \underline{ 0.2}& {\bf 6.0} & {\bf 0.3} &  5.0 \\
					& 20 & 6.6 & 0.6 & 9.7 & 1.0& {\bf 6.1} & {\bf 0.4}& \underline{6.2} & \underline{0.4} &  1.0 \\
					& 30 & 9.1 & 0.9 & 18.3 & 1.5& \underline{7.9} & \underline{1.0}& {\bf 7.5} & {\bf 1.2} &  0.5 \\
					\hline
					
					Parkinsons
					& 0 & 10.5 & 4.0& {\bf 7.1} & {\bf 3.3}* & 10.2 & 3.4& \underline{9.3} & \underline{3.6} &  5.0 \\
					& 10 & 13.8 & 4.2 & 12.9 & 4.0& {\bf 12.2} & {\bf 3.7}& \underline{12.7} & \underline{4.2} &  1.5 \\
					& 20 & 18.6 & 5.7 & 20.1 & 5.9& {\bf 16.2} & {\bf 4.8}& \underline{17.0} & \underline{4.9} &  0.75 \\
					& 30 & 24.1 & 5.5 & 28.0 & 6.0& \underline{23.0} & \underline{5.8}& {\bf 22.6} & {\bf 6.1} &  0.5 \\
					\hline
					Pima
					& 0 & 23.9 & 1.9 & 25.9 & 1.9& {\bf 23.2} & {\bf 2.0}& \underline{23.4} & \underline{1.8} &  0.5 \\
					& 10 & 25.9 & 2.3 & 28.9 & 2.3& \underline{24.7} & \underline{2.2}& {\bf 24.1} & {\bf 2.4} &  0.25 \\
					& 20 & 28.4 & 2.6 & 32.9 & 3.3& \underline{26.7} & \underline{2.5}& {\bf 25.3} & {\bf 2.5}* &  0.25 \\
					& 30 & 32.6 & 3.0 & 38.4 & 3.3& \underline{31.5} & \underline{2.8}& {\bf 29.8} & {\bf 3.7}* &  0.5 \\
					\hline
					
					Ringnorm
					& 0 & 8.9 & 1.8& {\bf 4.3} & {\bf 0.6}* & 6.0 & 1.1& \underline{4.4} & \underline{0.8} &  20.0 \\
					& 10 & 9.6 & 1.7 & 7.6 & 1.0& \underline{6.7} & \underline{1.2}& {\bf 6.1} & {\bf 1.4}* &  10.0 \\
					& 20 & 10.8 & 1.8 & 13.0 & 1.7& {\bf 7.9} & {\bf 1.4}*& \underline{8.4} & \underline{1.8} &  1.25 \\
					& 30 & 15.6 & 2.9 & 22.2 & 2.3& {\bf 12.4} & {\bf 2.4}& \underline{12.5} & \underline{3.0} &  0.75 \\
					\hline
					Sonar
					& 0 & 22.4 & 5.1& {\bf 15.1} & {\bf 4.6} & 17.9 & 4.8& \underline{15.5} & \underline{4.5} &  10.0 \\
					& 10 & 23.3 & 5.5 & 20.7 & 5.1& \underline{20.6} & \underline{5.4}& {\bf 19.7} & {\bf 4.6} &  5.0 \\
					& 20 & 26.9 & 5.7 & 26.3 & 5.3& {\bf 24.2} & {\bf 5.7}& \underline{24.5} & \underline{5.6} &  1.25 \\
					& 30& 32.2 & 5.0 & 34.6 & 5.5& {\bf 30.4} & {\bf 5.4}& \underline{ 30.4} & \underline{ 5.3} &  0.75 \\
					\hline
					Spambase
					& 0 & 5.9 & 0.5& \underline{4.3} & \underline{0.4} & 5.0 & 0.5& {\bf 4.1} & {\bf 0.5}& 20.0\\
					& 10 & 8.2 & 0.8& {\bf 6.1} & {\bf 0.6} & 6.5 & 0.6& \underline{6.3} & \underline{0.6}&0.75\\
					& 20 & 11.5 & 1.0& \underline{7.5} & \underline{0.7} & 9.4 & 0.8& {\bf 7.1} & {\bf 0.7}&0.25\\
					& 30 & 16.5 & 1.1& \underline{10.3} & \underline{1.1} & 14.3 & 1.0& {\bf 8.6} & {\bf 0.8}*&0.25\\
					\hline
				\end{tabular}}
				\end{table}
	\begin{table}
		
		\caption{Test error rates for bagging (unpruned CART trees), boosting (pruned CART trees),
		random forest, and vote-boosting (random trees).}
		\label{tab:results3}
		\centering \resizebox{\linewidth}{!}{%
			
			\begin{tabular}
				{@{}l@{$\;$}|r|r@{$\pm$}l@{$\;$}|r@{$\pm$}l@{$\;$}|r@{$\pm$}l@{$\;$}|r@{$\pm$}l@{$\;$}|p{1.1cm}c@{}}
				\hline
				& Noise   \\
				Dataset  & (in \%) &
				\multicolumn{2}{c|}{Bagging} &
				\multicolumn{2}{c|}{AdaBoost} &
				\multicolumn{2}{c|}{Random forest} &
				\multicolumn{2}{c|}{Vote-boosting} &
				\multicolumn{2}{c}{a=b (median)}  \\ 
				\hline
				Threenorm
				& 0 & 19.0 & 1.7 & 16.8 & 0.8& \underline{16.6} & \underline{0.9}& {\bf 16.2} & {\bf 1.0}* &  5.0 \\
				& 10 & 20.6 & 1.7 & 20.2 & 1.2& {\bf 18.5} & {\bf 1.1}& \underline{18.6} & \underline{1.2} &  1.5 \\
				& 20 & 23.3 & 1.8 & 24.9 & 1.6& {\bf 21.2} & {\bf 1.3}*& \underline{21.6} & \underline{1.5} &  1.25 \\
				& 30 & 28.8 & 2.1 & 32.0 & 1.9& {\bf 26.9} & {\bf 2.0}*& \underline{27.2} & \underline{2.5} &  0.62 \\
				\hline	
				Tic-tac-toe
				& 0 & 2.1 & 0.9& {\bf 0.7} & {\bf 0.6}* & 2.1 & 1.0& \underline{1.1} & \underline{0.7} &  10.0 \\
				& 10& \underline{5.7} & \underline{1.6} & 9.9 & 1.8 & 5.9 & 1.6& {\bf 5.6} & {\bf 1.6} &  2.5 \\
				& 20& {\bf 12.5} & {\bf 2.2} & 20.6 & 2.7& \underline{12.7} & \underline{2.6} & 13.0 & 2.6 &  1.25 \\
				& 30 & 23.5 & 2.8 & 30.3 & 2.9& {\bf 22.2} & {\bf 2.7}& \underline{22.9} & \underline{2.9} &  0.75 \\
				\hline
				Twonorm
				& 0 & 6.4 & 1.5 & 4.0 & 0.5& \underline{3.9} & \underline{0.5}& {\bf 3.6} & {\bf 0.5}* &  5.0 \\
				& 10 & 7.3 & 1.6 & 7.1 & 0.9& {\bf 4.8} & {\bf 0.7}*& \underline{4.9} & \underline{0.8} &  1.25 \\
				& 20 & 9.3 & 2.1 & 12.6 & 1.6& {\bf 6.6} & {\bf 1.1}& \underline{6.7} & \underline{1.2} &  0.75 \\
				& 30 & 14.2 & 2.3 & 21.5 & 2.4& \underline{10.7} & \underline{1.7}& {\bf 9.6} & {\bf 2.5}* &  0.5 \\
				\hline
				{\bf win/draw/loss} & 0 & 
				\multicolumn{2}{c|} { 12/11/0} &\multicolumn{2}{c|} { 6/13/4} & \multicolumn{2}{c|}{ 9/14/0} &  \multicolumn{2}{c|}{---}\\
				& 10 & 
				\multicolumn{2}{c|} { 16/7/0} &\multicolumn{2}{c|} { 17/6/0} & \multicolumn{2}{c|}{ 4/18/1} &  \multicolumn{2}{c|}{---}\\
				& 20 & 
				\multicolumn{2}{c|} { 18/5/0} &\multicolumn{2}{c|} { 16/7/0} & \multicolumn{2}{c|}{ 7/14/2} &  \multicolumn{2}{c|}{---}\\
				& 30 & 
				\multicolumn{2}{c|} { 19/4/0} &\multicolumn{2}{c|} { 19/2/2} & \multicolumn{2}{c|}{ 11/11/1} &  \multicolumn{2}{c|}{---}\\
			
					\hline
				\end{tabular}}
			\end{table}

\begin{figure*}[!t]
	\centering
\includegraphics[scale=0.475]{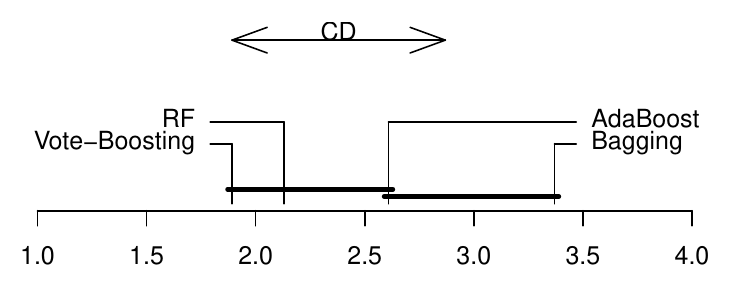}
\includegraphics[scale=0.475]{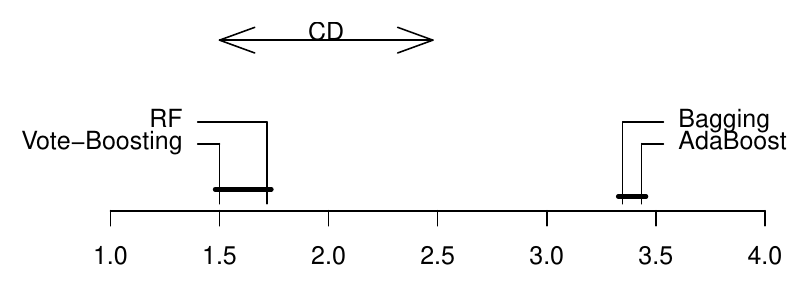}
\includegraphics[scale=0.475]{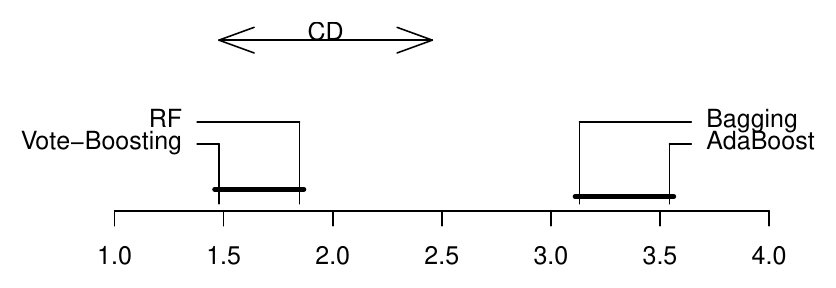}
\includegraphics[scale=0.475]{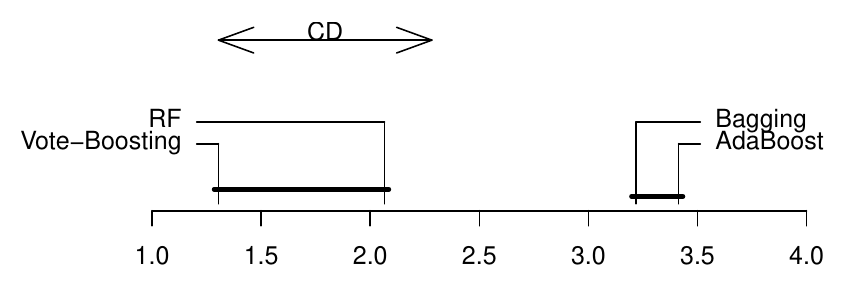}
  \caption{Comparison of the average ranks of bagging, AdaBoost, random forest
and vote-boosting using a Nemenyi test. Horizontal lines connect methods whose
average ranks are not significantly different (p-value $<$ 0.05). The 
plots correspond to datasets without injected noise (top left), with $10\%$
 (top right), $20\%$ (bottom left), and $30\%$ class-label noise (bottom right).} \label{demsar}
\end{figure*}

The test error rates of the different ensembles and classification problems 
considered are displayed in Tables~\ref{tab:results1}, \ref{tab:results2},
and \ref{tab:results3}. The results reported are averages, followed by the
corresponding standard deviations after the $\pm$ sign, over $100$
realizations of the training and test set partitions.  
The median value for $a=b$ used in vote-boosting is reported in the
last columns of Tables~\ref{tab:results1},
\ref{tab:results2}, and \ref{tab:results3}. In these tables, the best and  
second best results for each classification problem are highlighted
using bold face and underlined, respectively. 
In addition, the lowest test error rate is marked with an asterisk (*) if the
improvement over to the second best is statistically significant,
at a significance level $\alpha =0.05$.
The significance of these differences is determined using a paired resampled
t-test for synthetic problems, and to a
corrected resampled paired t-test \cite{nadeau+bengio_2003_inference,bouckaert+frank_2004_evaluating}
when random train/test partitions are used. Finally,
the number of significant wins and losses
when one compares the average accuracy of
vote-boosting with each of the remaining ensembles, 
according to the aforementioned statistical test,
 are shown in the last row of
Table~\ref{tab:results3}. 
Draws correspond to differences of average accuracy
that are not statistically significant.

From the results presented in these tables, it is clear that
vote-boosting ensembles exhibit the best overall performance.
In particular, it has the largest number of statistically significant wins 
at all noise levels. The tally is very favorable when
one compares the average accuracy of vote-boosting with bagging: vote-boosting
significantly outperforms bagging in $12$ out of the $23$ datasets (without
injected noise). For 10\%, 20\% and 30\% noise levels, the differences in number
of wins become larger: vote-boosting significantly outperforms bagging in
$16$, $18$ and $19$ out of the $23$ tested datasets, respectively.
The comparison with AdaBoost on the explored datasets is also
favorable to vote-boosting. In the original datasets (i.e.without injected
noise) vote-boosting outperforms AdaBoost in $6$ out
of $23$ datasets and is inferior in $4$ datasets: 
{\it Ozone}, {\it Parkinsons}, {\it Ringnorm}, and {\it Tic-tac-toe}).
In this problems vote-boosting is second best; furthermore, its 
test error rates of vote-boosting are fairly close
to AdaBoost. In these classification problems, 
except for {\it Ozone}, the shape
parameter of the beta distribution used as emphasis function in vote-boosting is
fairly high.
This indicates a strong emphasis on uncertain examples, which has a similar
effect as the emphasis on incorrectly classified instances that is
characteristic of AdaBoost. On the other hand, in problems such as {\it Blood},
{\it Heart}, {\it Liver} or {\it Pima}, which are difficult for AdaBoost,
vote-boosting selects low values for $a=b$, which implies that less emphasis
is made on uncertain examples.

It is remarkable that, for some datasets, 
such as {\it Blood}, {\it Heart}, and {\it Pima},
and in most of the problems, for sufficiently high
levels of class-label noise,
values of $a = b$ below $1.0$ provide the best accuracy. 
In such cases, emphasis is made on
instances on which the individual ensemble classifiers agree the most.
The effectiveness of this type of emphasis, which is somewhat counter-intuitive, is
a consequence of the large intrinsic variability of random trees. In fact, this
effect is less prominent in ensembles of more stable base learners, 
such as decision stumps or pruned CART trees. 
As the noise level increases, the performance of AdaBoost
rapidly deteriorates. By contrasnt vote-boosting is robust to class-label
noise: it outperforms AdaBoost in $17$, $16$ and
$19$ datasets for 10\%, 20\% and 30\% injected noise, 
respectively.

Finally, vote-boosting is more accurate than random forest in $9$ out of the
$23$ classification problems investigated for the noiseless case. 
In some cases, the improvements can be fairly
large (e.g. {\it Blood}, {\it Chess}, {\it Ringnorm}, {\it Sonar} or {\it
Tic-tac-toe}).
In $14$ datasets the two methods have comparable accuracies. 
The accuracy improvements of random forests over vote-boosting
are not statistically significant in any of the problems
investigated.
When the class labels are contaminated, 
vote-boosting performs significantly better in
$4$, $7$ and $11$, and worse in $1$, $2$ and $1$ datasets for 10\%, 20\%, and
30\% noise levels, respectively. 
Again, in the noisy datasets we see that, when
random forest wins, the differences are typically small. By 
contrast, when vote-boosting wins, the differences are, in general, large. 

In addition, the overall performance of the accuracies of the different ensembles
are summarized in \figurename~\ref{demsar} using the procedure described in 
\cite{demsar_2006_statistical}.
The plots in this figure display the average ranks of bagging,
AdaBoost, random forest (RF) and vote-boosting for the original datasets (top
left), and for 10\% (top right), 20\% (bottom left), 30\% (bottom right)
injected class-label noise. In these plots, 
a horizontal solid line connects
methods for which the differences of
average ranks are not statistically significant 
using a Nemenyi test with p-value $<$ 0.05.
In the original problems, vote-boosting has the best average rank. However, the
differences with random forest and AdaBoost are not statistically significant.
The difference with bagging, which has the worst performance in terms average
rank, is statistically significant.  For problems contaminated with noise in the
class labels, vote-boosting has the best average rank. The differences with
respect to random forest increase for higher noise levels. However, they are not
statistically significant. In all noisy problems, the average ranks of both
random forest and vote-boosting ensembles are significantly better than bagging
and AdaBoost.

\begin{figure*}
\centering
\includegraphics[scale=0.5]{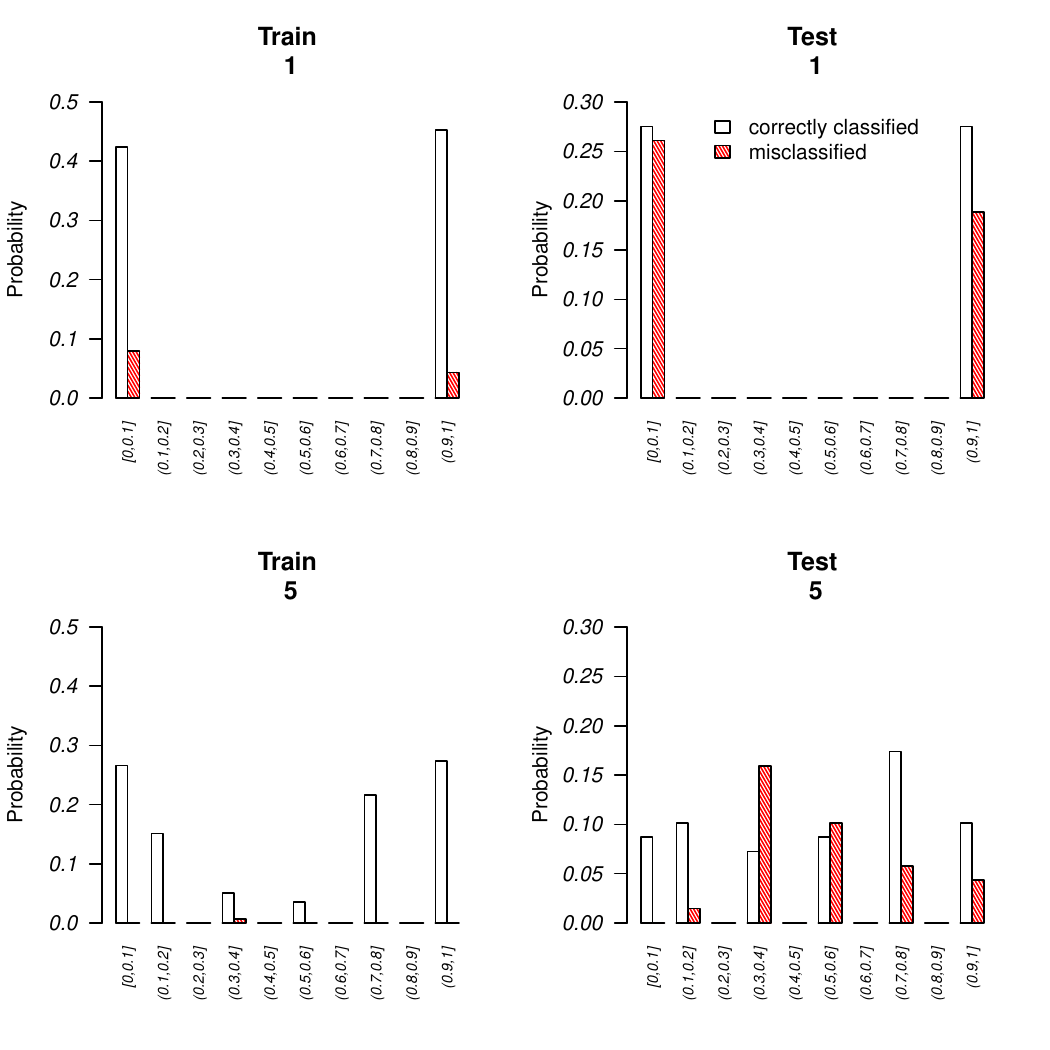}
\includegraphics[scale=0.5]{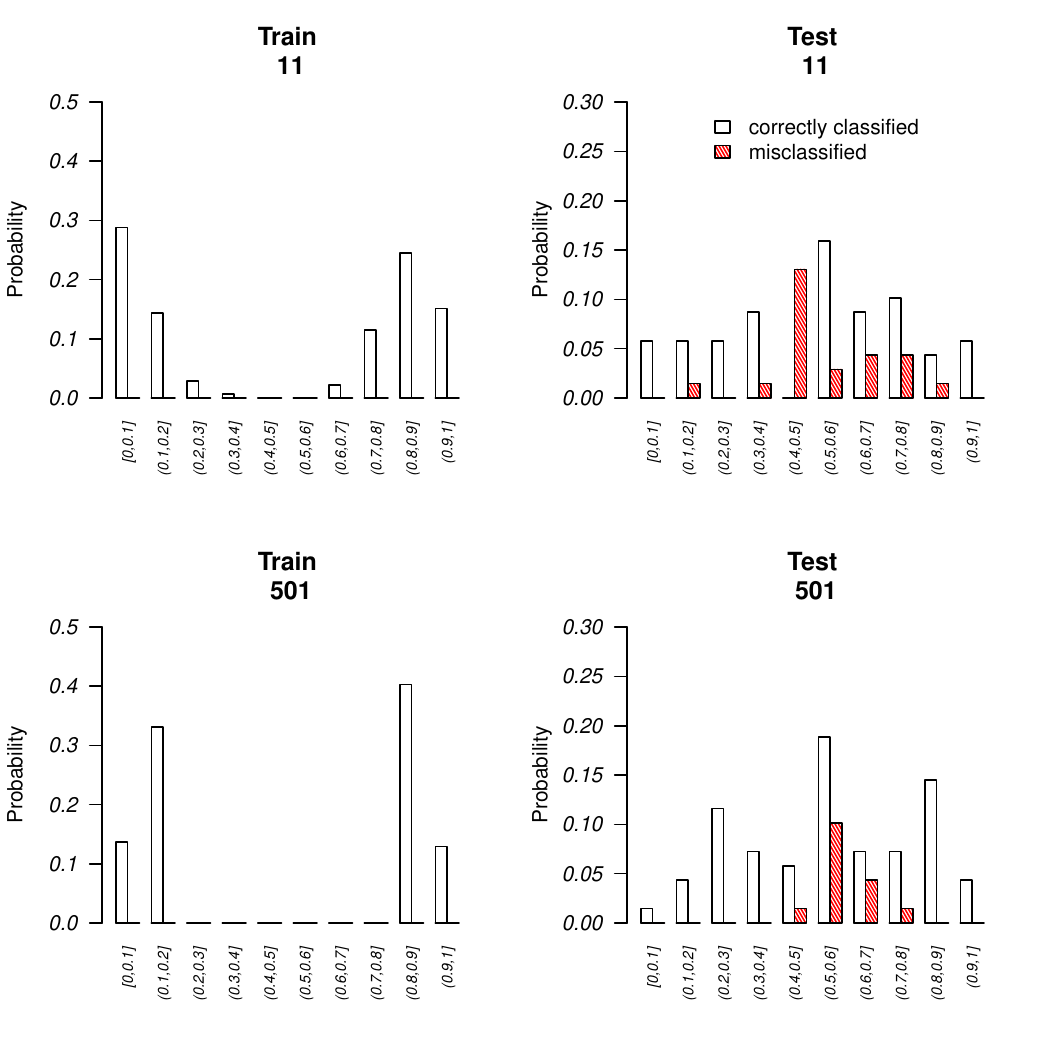}
\caption{
Histograms of vote fractions for correctly (white) and incorrectly (red)
classified instances in {\it Sonar} for the training set (left column) and test
set (right column)}
\label{hist1}
\end{figure*}

\begin{figure*}
\centering
\includegraphics[scale=0.5]{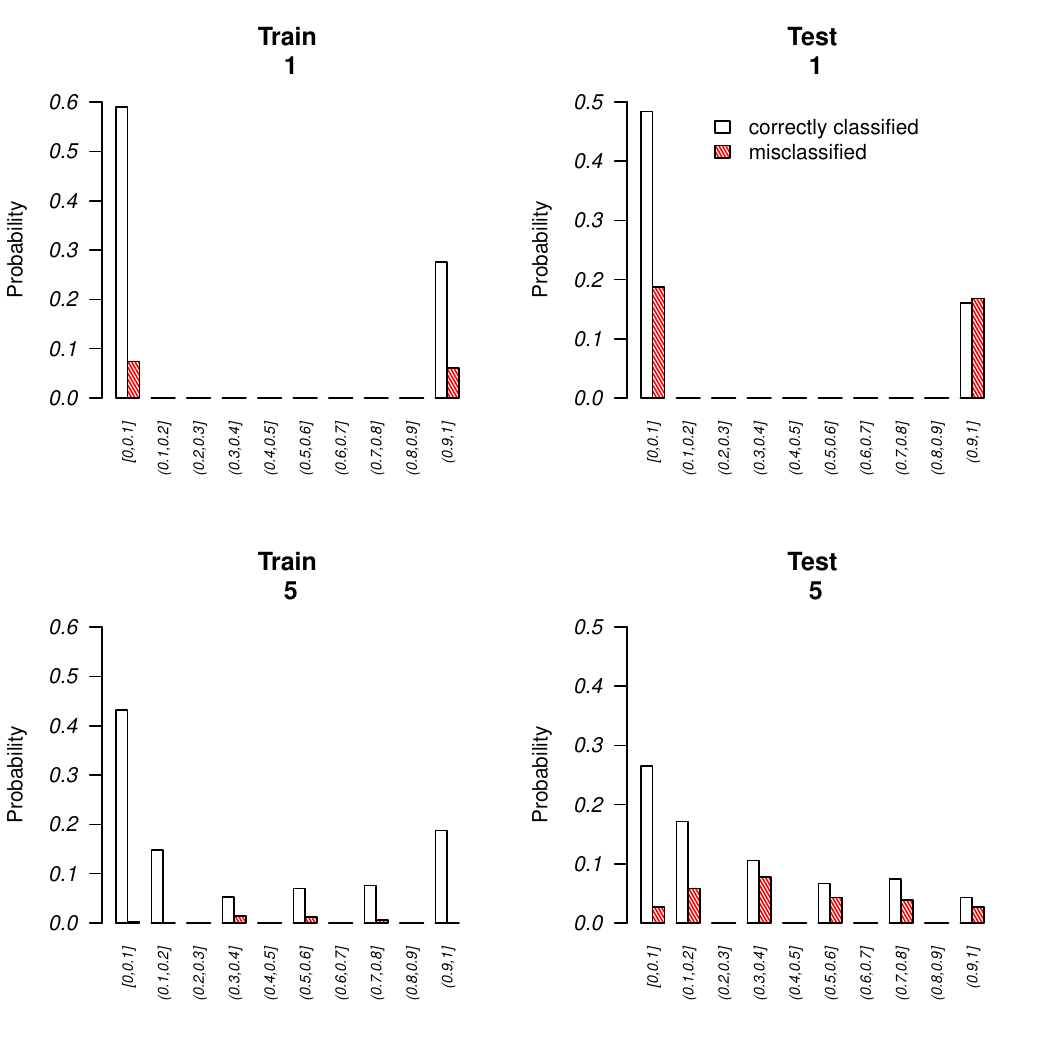}
\includegraphics[scale=0.5]{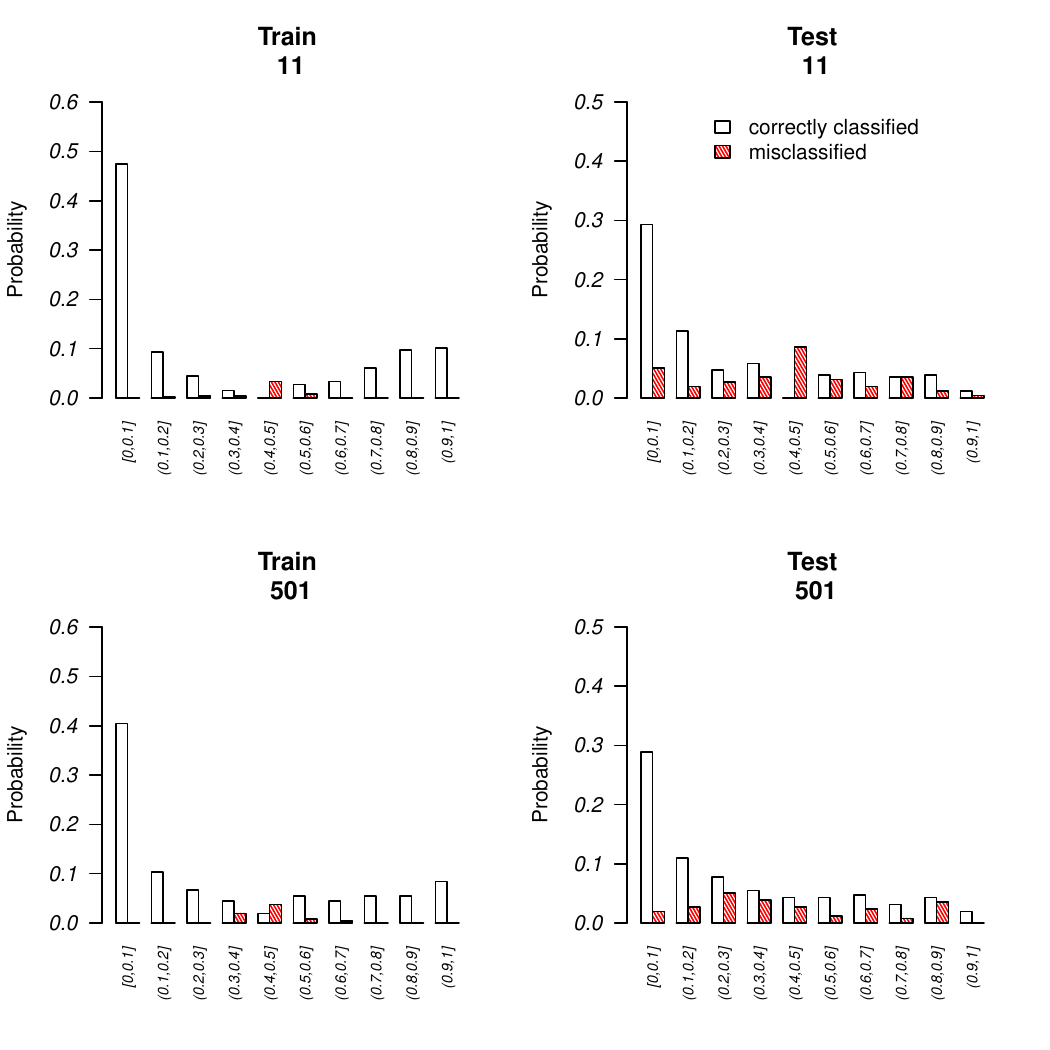}
\caption{
Histograms of vote fractions for correctly (white) and incorrectly (red)
classified instances in {\it Pima} for the training set (left column) and test
set (right column)}
\label{hist2}
\end{figure*}

\subsection{Emphasis profiles}

From the values of the shape parameter of the beta distribution reported in 
the last column of Tables~\ref{tab:results1}, \ref{tab:results2}, and \ref{tab:results3}, 
it is apparent that the type of emphasis and its strength need
to be adapted to the classification task considered. To further investigate 
this point, we have carried out a detailed analysis of the distribution of class votes
for {\it Sonar} and {\it Pima}, in which the optimal emphasis strategies are very
different.

In {\it Sonar}, the vote-boosting ensemble is built using a beta 
distribution with $a = b = 10.0$. Thus, the optimal emphasis is to 
focus on training instances in which the disagreement rates are largest. 
In \figurename~\ref{hist1}, 
the histograms of the distribution of votes are 
plotted for ensembles of $1$, $5$, $11$ and $501$ random trees. 
The height of the white bars indicate the fraction of correctly classified
instances for the corresponding range of voting distributions. 
The red stripped bars correspond to incorrectly classified instances. 
The plots on the left are for the training set and on the right for the test 
set. In the training set, the strong focus on uncertain instances 
(those for which the fraction of class votes is close to $0.5$) leads 
to a markedly bimodal distribution, in which most predictions are by clear majority. 
Incorrectly classified instances disappear because ensembles that are
sufficiently large achieve zero training error. 
The distribution of class votes in the test set is markedly
different: It covers the whole interval, and exhibits a low peak for
intermediate class vote frequencies, especially for instances that are
misclassified.

A very different picture is obtained in {\it Pima} (\figurename~\ref{hist2}). In
this classification task, the selected shape parameter for the symmetric beta
distribution is $ a = b = 0.5$. In consequence, the optimal strategy is to avoid
focusing on training instances in which the disagreement rates are large. For
correctly classified instances, the histograms in training and test
sets are similar. Misclassified
instances in the training set appear mostly around $0.5$. By contrast, in the test
set, they appear in the whole $[0,1]$ interval. This is consistent with 
the observation that {\it Pima} has high levels of class-label noise.

\section{Conclusions} \label{sec:conclusions}

Vote-boosting is a novel ensemble learning method in which individual
classifiers are built using different weighted versions of the training
data. To build a new classifier, the weights of the training
instances are determined 
in terms of the disagreement rate among the classifiers that 
make up the ensemble at that point. 
The optimal weighing scheme depends on the complexity of the base classifiers
and on the level of noise in the class labels of the training data.  
For simple or regularized classifiers, such
as decision stumps or pruned CART trees, vote-boosting interpolates between
bagging and AdaBoost. When the level of class-label noise is small, prediction
errors are more likely to occur near the classification boundary, 
where the uncertainty, as measured by the 
disagreement among ensemble predictions, is largest. Therefore, it
is possible to build more accurate ensembles by focusing on uncertain instances.
Since these instances are more likely to be misclassified, the emphasis given by
vote-boosting is similar to AdaBoost's. For noisy classification problems, a
softer emphasis on uncertain instances is generally preferable. In this case,
the most accurate predictions are obtained by means of ensembles that are fairly
similar to bagging.
When more brittle individual learners are used (e.g. unpruned CART or
random trees) a milder emphasis on uncertain instances 
is generally needed to achieve the best
generalization performance. This is a consequence of the fact that some of the
uncertainty in the predictions is due to the intrinsic variability of the base
learners. For problems in which the level of class-label noise is  high
it is in fact advantageous to progressively focus on instances in which the
ensemble classifiers agree. In practice, 
the optimal type of emphasis can be readily
determined using cross-validation within the training data.

Note that, since AdaBoost is based on emphasizing incorrectly classified
instances, it cannot be used to improve the performance of base learners whose
training error is small, such as unpruned CART or random trees. 
By contrast, vote-boosting does
not have this limitation and can be used to build boosted ensembles composed
of these types of classifiers that are both accurate and robust to noise
in the class labels.

			
%

			
%
			
\section*{Acknowledgements}

The research has been supported by the Spanish {\it Ministry of Economy, Industry,
and Competitiveness} (projects TIN2016-76406-P, TIN2013-42351-P and
TIN2015-70308-REDT), and {\it Comunidad de Madrid}, 
project CASI-CAM-CM (S2013/ICE-2845). 


\bibliographystyle{elsarticle-harv}
\bibliography{vote_boosting}
			
			
			
			
			

			
\end{document}